\documentclass[lettersize,journal]{IEEEtran}
\usepackage{amsmath,amsfonts}
\usepackage{algorithmic}
\usepackage{algorithm}
\usepackage{array}
\usepackage[caption=false,font=normalsize,labelfont=sf,textfont=sf]{subfig}
\usepackage{textcomp}
\usepackage{stfloats}
\usepackage{url}
\usepackage{verbatim}
\usepackage{graphicx}
\usepackage{cite}
\usepackage[usenames, dvipsnames]{color} 
\definecolor{purple}{rgb}{0.5, 0.0, 0.5}
\definecolor{orange}{rgb}{1, 0.65, 0}
\definecolor{lightgreen}{rgb}{0.68, 1, 0.18}
\definecolor{darkgreen}{rgb}{0.09, 0.32, 0.24}
\definecolor{darkred}{rgb}{0.6, 0, 0}
\definecolor{brown}{rgb}{0.64, 0.16, 0.16}
\definecolor{darkpink}{rgb}{1, 0.08, 0.57}

\usepackage{hyperref} 
\usepackage{comment} 
\usepackage[table]{xcolor} 

\usepackage{tabularx,booktabs} 
\usepackage{amssymb,pifont} 
\usepackage{anyfontsize}
\usepackage{multirow}
\usepackage{enumitem} 
\usepackage{rotating} 

\newcommand{\cmark}{\ding{51}}%
\newcommand{\xmark}{\ding{53}}%
\usepackage{tablefootnote}

\iffalse
  \newcommand{\venice}[1]{\noindent}
  \newcommand{\holger}[1]{\noindent}
  \newcommand{\wk}[1]{\noindent}
  \newcommand{\ks}[1]{\noindent}
  \newcommand{\done}[1]{\noindent}
  \newcommand{\todo}[1]{\noindent}
\else
  \newcommand{\wk}[1]{\textcolor{blue}{\bf [WF: #1]}}
  \newcommand{\holger}[1]{\textcolor{purple}{\bf [HC: #1]}}
  \newcommand{\venice}[1]{\textcolor{brown}{\bf [VL: #1]}}
  \newcommand{\ks}[1]{\textcolor{darkpink}{\bf [KS: #1]}}
  \newcommand{\done}[1]{\textcolor{darkgreen}{\bf [Done: #1]}}
  \newcommand{\todo}[1]{\textcolor{red}{\bf [Todo: #1]}}
\fi

\begin{document}

\title{nuScenes Revisited: Progress and Challenges in Autonomous Driving}

\author{
Whye Kit Fong,
Venice Erin Liong, 
Kok Seang Tan,
Holger Caesar
\IEEEcompsocitemizethanks
{\IEEEcompsocthanksitem
Corresponding author: whyekit.fong@motional.com
}
\thanks{Manuscript received Mm DD, 2025.}}

\markboth{Journal of \LaTeX\ Class Files,~Vol.~14, No.~8, August~2021}%
{Shell \MakeLowercase{\textit{et al.}}: A Sample Article Using IEEEtran.cls for IEEE Journals}

\IEEEpubid{0000--0000/00\$00.00~\copyright~2021 IEEE}

\maketitle

\begin{abstract}
Autonomous Vehicles (AV) and Advanced Driver Assistance Systems (ADAS) have been revolutionized by Deep Learning. 
As a data-driven approach, Deep Learning relies on vast amounts of driving data, typically labeled in great detail.
As a result, datasets, alongside hardware and algorithms, are foundational building blocks for the development of AVs.
In this work we revisit one of the most widely used autonomous driving datasets: the nuScenes dataset.
nuScenes exemplifies key trends in AV development, being the first dataset to include radar data, to feature diverse urban driving scenes from two continents, and to be collected using a fully autonomous vehicle operating on public roads, while also promoting multi-modal sensor fusion, standardized benchmarks, and a broad range of tasks including perception, localization \& mapping, prediction and planning.
We provide an unprecedented look into the creation of nuScenes, as well as its extensions nuImages and Panoptic nuScenes, summarizing many technical details that have hitherto not been revealed in academic publications. 
Furthermore, we trace how the influence of nuScenes impacted a large number of other datasets that were released later and how it defined numerous standards that are used by the community to this day.
Finally, we present an overview of both official and unofficial tasks using the nuScenes dataset and review major methodological developments, thereby offering a comprehensive survey of the autonomous driving literature, with a particular focus on nuScenes.
\end{abstract}

\begin{IEEEkeywords}
Autonomous driving, dataset, perception, localization, mapping, prediction, planning
\end{IEEEkeywords}

\vspace{30pt}

\IEEEraisesectionheading{\section{Introduction}}
\label{sec:introduction}
\IEEEPARstart{I}{n} the beginning there was KITTI~\cite{geiger2012kitti}.
The KITTI dataset was the first large-scale multimodal dataset for autonomous driving (AD) with 3D object annotations.
Following major milestones in autonomous driving, such as the release of the Velodyne HDL-64E lidar sensor (2007) and the launch of the Google self-driving car project (2009), the KITTI dataset (2012) first made multi-sensor AD data available to a broad community, facilitating research into stereo vision, optical flow, depth estimation, odometry, object detection and tracking, as well as various segmentation tasks.
However, KITTI's constraints in terms of scale, sensor configuration, and environmental diversity make it inadequate for commercial robotaxi applications.

In 2016 nuTonomy (later called Motional), a company founded by MIT researchers, offered the world's first self-driving taxi service as a pilot program in One North, Singapore. 
The same vehicles were later used to collect the ``nuTonomy 1000 scenes dataset``, short nuScenes$^{T\kern-1.35pt M}$~\cite{caesar2020nuscenes}, with an initial release in 2018 and a full release in 2019.
The dataset reflects the environments and challenges encountered by nuTonomy's self-driving taxis during their operations, and it serves as a benchmark for evaluating and comparing autonomous driving models.
nuScenes is designed to address the limitations of KITTI and make it viable sandbox for industry and academia to research autonomous vehicles.
In particular, it allows for free use by academics, and provides licenses for commercial partners. 
It was the first dataset to provide a complete multimodal sensor suite with 360° camera coverage, lidar and radar sensors. Due to the robotaxi use-case (SAE Levels 4-5), the driving regions are mapped in advance, which allows for accurate localization even in the presence of urban canyons with limited GNSS reception.
Furthermore, semantic maps simplify the interpretation of the environment and have proven to be essential for prediction and planning tasks. 
Lastly, the nuScenes dataset features more diverse environments than existing datasets, featuring data from two countries with different climates, vehicle types and driving directions. It is captured in dense urban environments and the broad class taxonomy allows users to focus particularly on rare classes.

In this paper we discuss the historical developments leading up to nuScenes and its influence on subsequent developments in the field.
We critically discuss good and bad design choices of the nuScenes dataset and give recommendations for future applications.
Finally, we focus on important benchmark tasks, rather than individual works and summarize the progress and challenges in autonomous driving as it relates to nuScenes. 

To summarize, our contributions are:%
\begin{itemize}[topsep=0pt, partopsep=0pt]
    \item An overview of the nuScenes dataset, with detailed explanations of the dataset creation process that are hitherto undocumented in academic works.
    \item A discussion of the pros and cons of the dataset, as well as how other datasets took inspiration from nuScenes and how they further improved upon it.
    \item An overview of popular autonomous driving tasks and their metrics, as well as a survey of leading methods on these tasks.
\end{itemize}

\IEEEpubidadjcol
\vspace{-2mm}
\section{Dataset}
\label{sec:dataset}

\begin{figure*}
  \includegraphics[width=\textwidth]{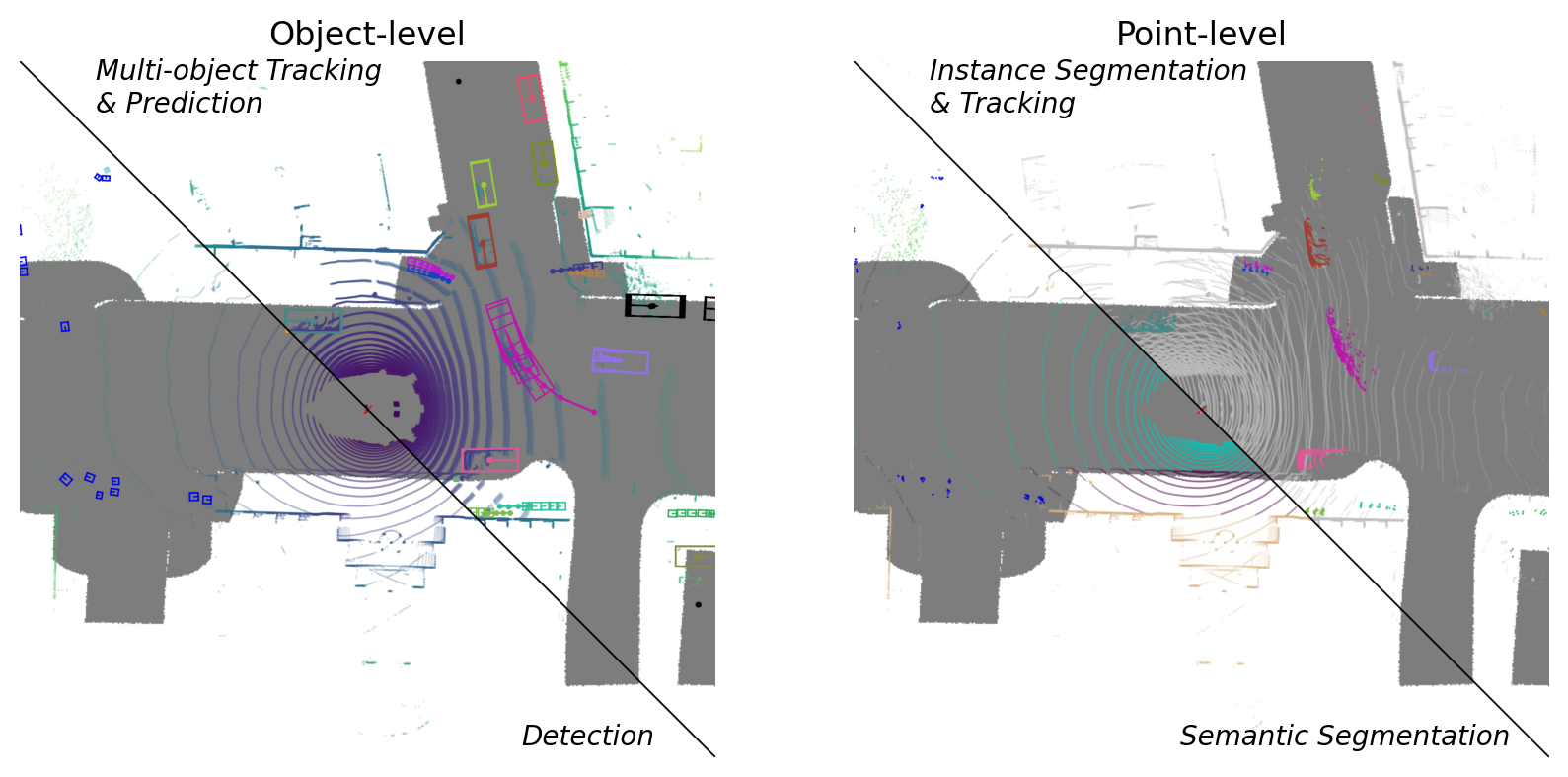}
  \caption{Representation of various perception tasks available in nuScenes. There are two types of perception tasks: 
  a.) \textbf{Object-level} tasks, where we detect, classify, track or predict various agents surrounding the ego represented by bounding-box instances and tracklets; 
  b.) \textbf{Point-level} tasks, where we segment agents by classifying individual points (e.g. lidar data).}
\end{figure*}

nuScenes is a large-scale multi-modal autonomous driving dataset which contains 1000 carefully selected driving scenes from two cities with left and right-hand driving.
The data collection vehicles are driven non-autonomously by expert drivers.
The dataset contains 3D bounding box annotations from 23 diverse object classes, and object-level attributes such as visibility, activity and pose. 

\subsection{Vehicle Setup}
\begin{figure}[ht]
\centering
\includegraphics[width=\linewidth]{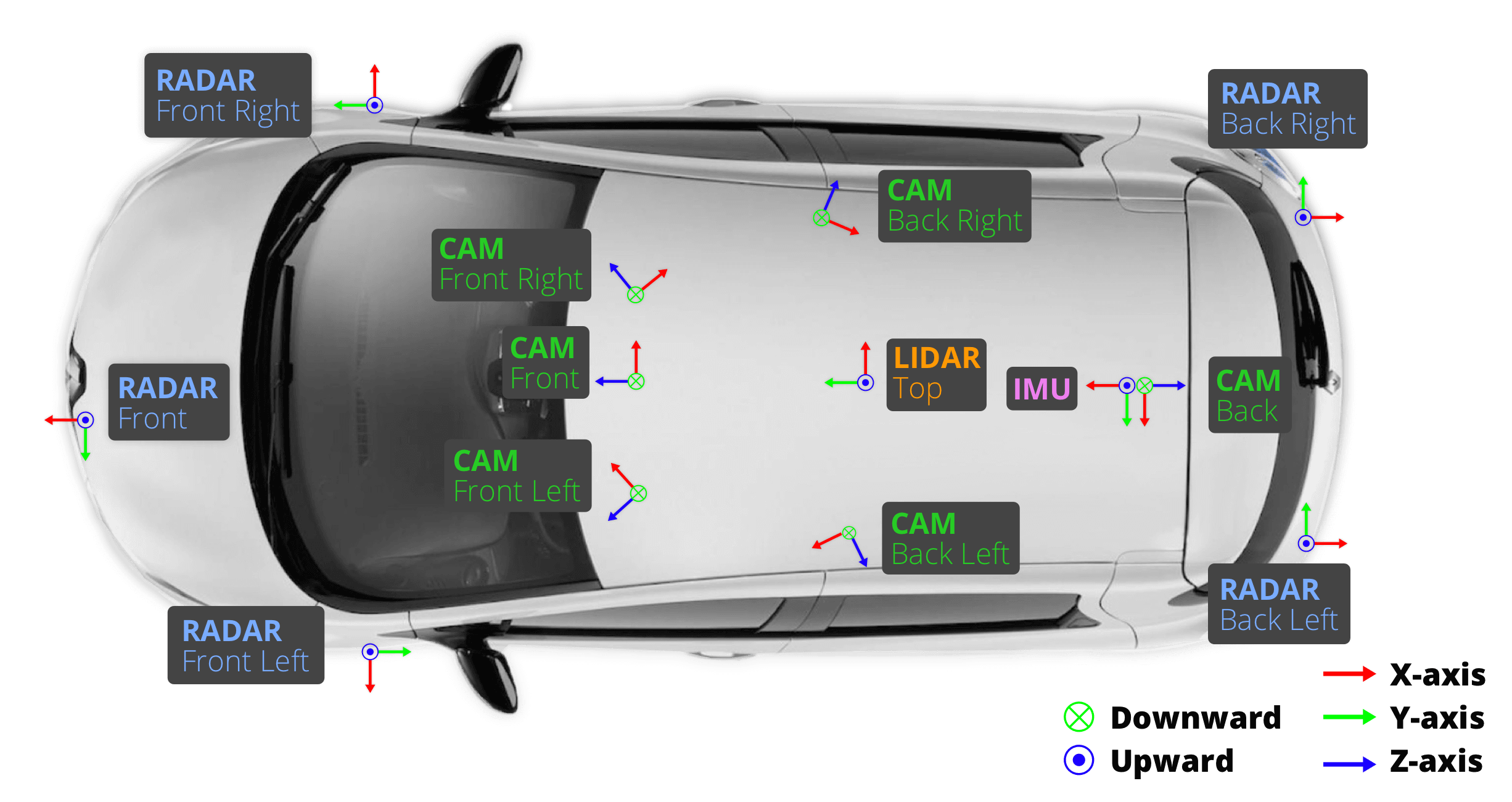}
\caption{Overview of the sensors and their positions on the nuScenes vehicles.
Figure taken from the original nuScenes publication~\cite{caesar2020nuscenes}.
}
\label{fig:ego_with_sensors}
\end{figure}

Two Renault Zoe supermini electric cars with an identical sensor layout were used for data collection in Boston and Singapore. 
Fig.~\ref{fig:ego_with_sensors} shows the sensor placement on the vehicles. 
The specific sensors used on the vehicle are:
\begin{itemize}
\item{Lidar: 1x Velodyne HDL32E}
\item{Camera: 6x Basler acA1600-60gc} 
\item{Radar: 5x Continental ARS 408-21}
\item{IMU \& GNSS: 1x Advanced Navigation Spatial}
\end{itemize}

Front and side cameras have a 70° FOV and are offset by 55°, while the rear camera has a FOV of 110°.
The focal length of the front and side cameras is 5.5 mm, while that of the rear camera is 4 mm. The sensor is the same across all cameras (1/1.8'' CMOS sensor of 1600x1200 resolution). The cameras are global shutter.
Data from additional blind spot lidars that are part of the vehicle setup were never released, due to data quality issues.
For more information on the sensors, refer to~\cite{caesar2020nuscenes} or the manufacturer's handbook.

\subsection{Data post-processing}
\label{sec:dataset-postprocessing}

\textbf{Image resolution.}
The camera images are cropped from 1600x1200 to 1600x900 by removing the top 300 pixels. This is done as the top portion of the vertical field-of-view is generally not relevant in the context of autonomous driving. In addition, this helps to save computational resources when processing the camera images.

\textbf{Ego-motion compensation.}
The lidar sweeps are ego-motion compensated to the ego position at the timestamp of the sweep themselves. 
This compensation is performed onboard the vehicle during data collection leveraging high-frequency odometry information.
It is baked into the dataset files and therefore cannot be undone (e.g. for research purposes).
Due to the ego-motion compensation and the use of a global shutter, we can convert annotations from the lidar to the camera frame (and vice versa) using a single transformation. 
This homogeneous transformation \( T \) depends on the position of the ego vehicle at both the lidar timestamp $t_l$ and the camera timestamp $t_c$, where:
\[ T = 
T_{(camera\leftarrow ego)} \; 
T_{(ego_{t_c}\leftarrow ego_{t_l})} \; T_{(ego\leftarrow lidar_)} \]
Here \( T_{(ego\leftarrow lidar)} \) refers to the transformation from the lidar frame to the ego frame, 
\( T_{(ego_{t_c}\leftarrow ego_{t_l})} \) refers to the transformation from the ego frame at time of lidar capture \( t_l \) to the ego frame at the image capture time \( t_c \), 
while \( T_{(camera\leftarrow ego)} \) refers to the transformation from the ego frame to the camera frame. 

\textbf{Synchronization.}
As detailed in~\cite{caesar2020nuscenes}, the cameras and the lidar in nuScenes are jointly triggered and timestamped and therefore well synchronized, while the radar does not support synchronization. 
The cameras operate at 12 Hz, while the lidar operates at 20 Hz. 
Thus, we sample one frame from each camera and the lidar at 2 Hz, to ensure that both sensors are well aligned. 
If no acceptable alignment can be achieved, we discard the entire scene. 
We visually inspect the quality of the synchronization by projecting lidar points of moving objects (relative to the ego vehicle) onto the corresponding camera images and verifying that object outlines are well aligned across camera and lidar.

\begin{figure} 
    \centering
    \includegraphics[width=\linewidth]{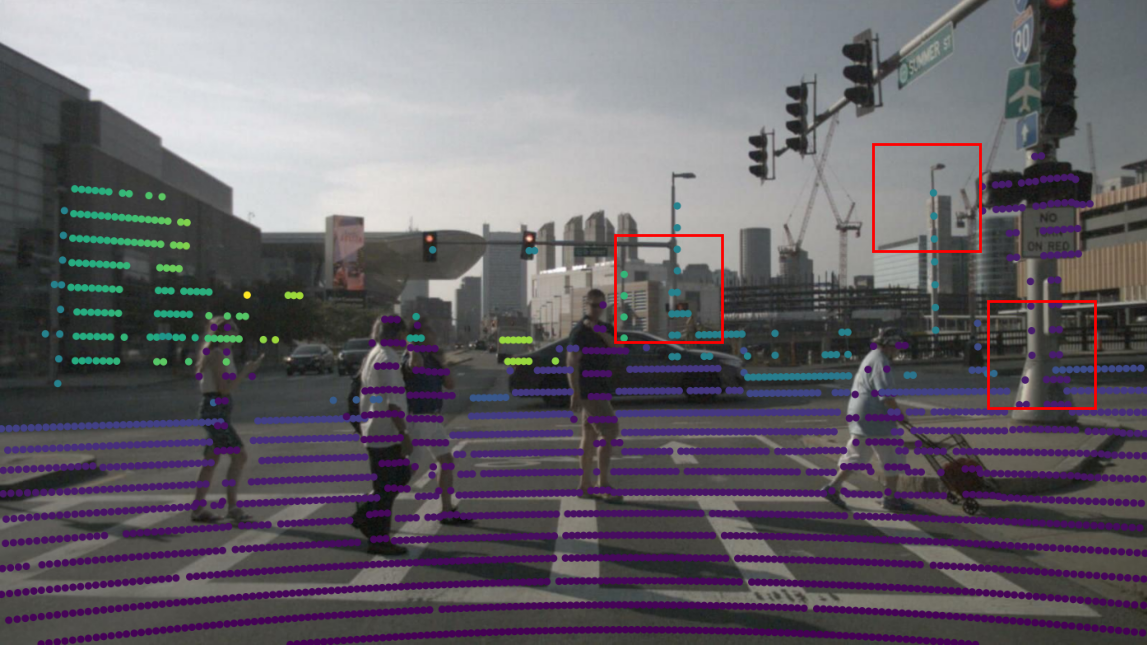}
    \\
  \subfloat[\label{one}]{%
       \includegraphics[width=0.25\linewidth]{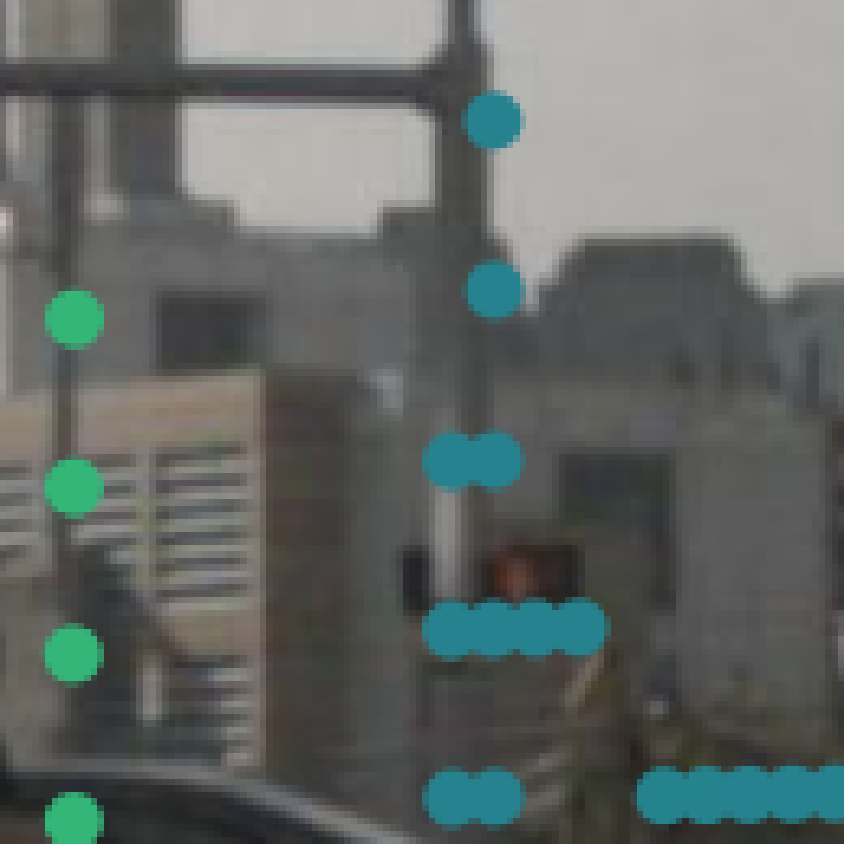}}
  \hspace{1em}
  \subfloat[\label{two}]{%
        \includegraphics[width=0.25\linewidth]{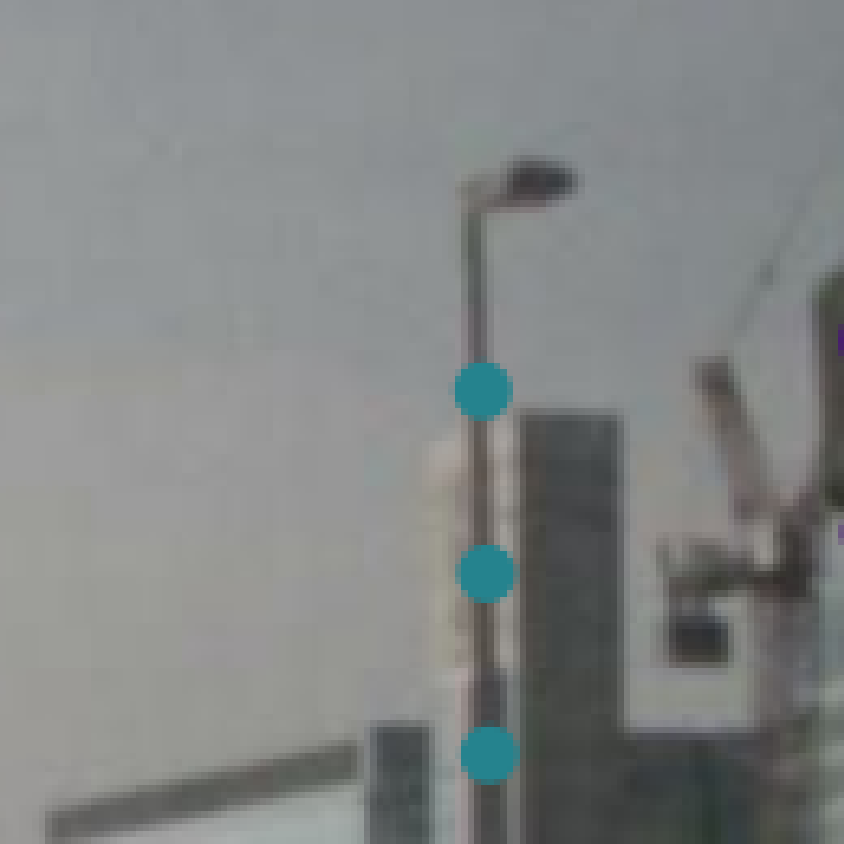}}
  \hspace{1em}
  \subfloat[\label{three}]{%
        \includegraphics[width=0.25\linewidth]{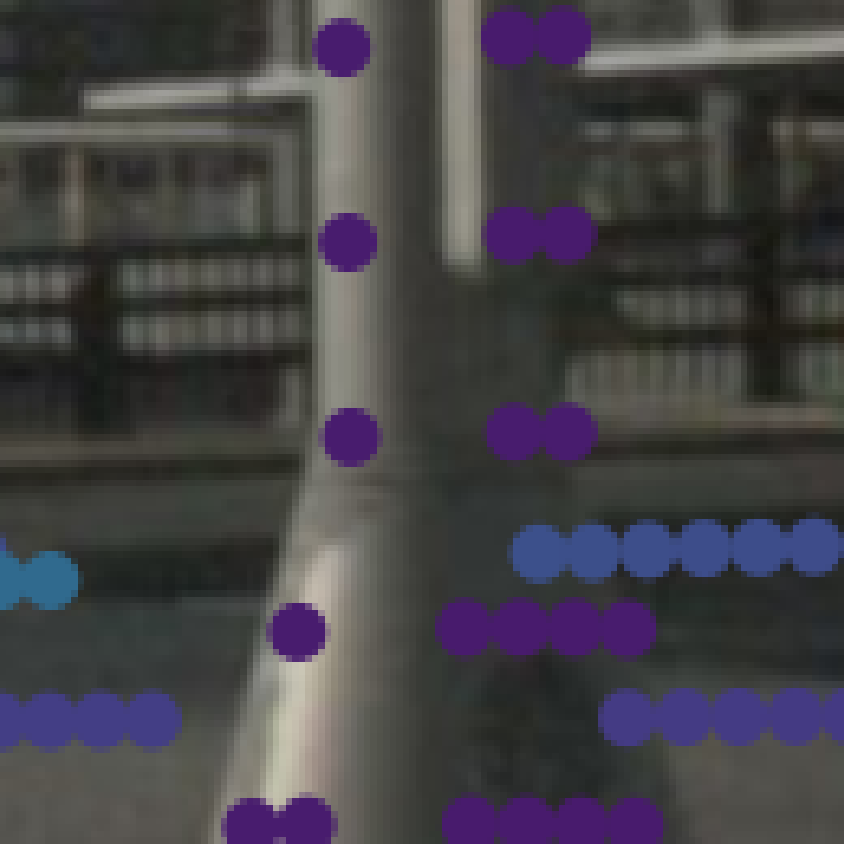}}
  \caption{Examples of the high quality calibration between camera and lidar in nuScenes.
  Apparent misalignments in (c) are due to the different locations of cameras and lidars (parallax).
  }
  \label{fig:camera_lidar_sync}
\end{figure}

\textbf{Calibration.}
As detailed in~\cite{caesar2020nuscenes}, all the sensors in the vehicles used in nuScenes are meticulously calibrated. 
Nevertheless, calibration can deteriorate over time and we make sure to check for each collected log that the calibration quality is still good.
To check the calibration, we follow a similar procedure to the synchronization, but look at points which fall on stationary objects (relative to the ego).
Fig.~\ref{fig:camera_lidar_sync} shows a qualitative example of the synchronization and calibration between a given camera and the lidar.
In case of systematic misalignments we use an in-house tool to visually adjust the calibration parameters to achieve a good fit between camera and lidar.

\subsection{Data Annotation}
As mentioned previously, the selected scenes are sampled at 2 Hz. These sampled frames are known as keyframes.
Due to the guaranteed synchronization of these keyframes, we can use both lidar and camera (and to a lesser extent radar) for the annotation.
The objects in each keyframe are annotated with a semantic category (one of the 23 object classes), attributes (visibility, activity and pose), an instance identifier, and a bounding box (represented by x, y, z, width, length, height and yaw angle).
A consistent taxonomy was used across nuScenes and its extensions~(see Section~\ref{sec:dataset_extensions}).
Object tracks are annotated continuously throughout each scene and identified by the instance identifier.
Bounding box interpolation is used to speed up the annotation process.
Boxes that do not contain any lidar or radar points are discarded, as they likely correspond to occluded or distant objects.
Using expert annotators and multiple validation steps, highly accurate annotations are achieved.
High annotation quality is ensured through expert annotators and multiple validation stages.
However, the lidar/radar-centric design introduces a bias: objects that are not visible in the camera but are annotated due to their presence in lidar or radar may lead to unfair penalties for camera-only perception methods, which are inherently unable to detect such objects.

\subsection{Privacy blurring}
To ensure the privacy of individuals captured in the nuScenes dataset, we blur regions in the images corresponding to faces and license plates. 
Inspired by the approach taken in Google Street View~\cite{frome2009streetview}, we develop an automated object detection and blurring system with a very high detection recall, to favor blurring too many regions over missing personally identifiable pedestrians.
This object detector is trained on publicly available datasets for person and license plate detection and we validate that it generalizes well to our data.
By projecting these boxes to the camera images, we can limit the area in which to apply the object detectors, since foreground classes (incl. pedestrians and vehicles) account for less than 7\% of all lidar points in nuScenes and cover a similar area in the camera images. 
This dramatically increases detection recall. 
Finally we apply a Gaussian image blurring filter with manually tuned parameters to the image region corresponding to the face or license plate.
For cases where our detector fails to detect faces or license plates or when the blurring is insufficient, we provide a privacy take-down form.
However, this is very rare given the high recall of the detector and the relatively low camera resolution.

\subsection{Quality assurance}
Quality assurance of the annotations is an essential part of dataset creation, yet it is rarely discussed for AV datasets. 
One reason for this is that annotations are typically outsourced to third-party providers that do not reveal their quality assurance protocol.
The same also applies to the nuScenes dataset, which was annotated by Scale AI, using multiple rounds of undocumented quality assurance. 
In addition, our team performed numerous manual and semi-automated checks.
The most time-consuming manual QA step was performed in the same multimodal 3D data viewer that was used to annotate the data. 
For every object track, we check the scale and class, which do not change over time.
In addition, for every cuboid in every track, we check the position, orientation and attributes, which are subject to change.
Since one can easily lose the overview in an interactive editor, we also render videos of each scene, visualizing the object bounding boxes, classes and attributes and thereby identifying several missing annotations.
After manual QA, we apply a number of semi-automated correction steps.
In a first step we enforce the orientation conventions of nuScenes.
For vehicles, we expect one of the shorter sides to be the front, and perform manual checks where this is not the case.
For barriers and cones, we define the front to be the (longer) side that faces the primary road that the ego vehicle drives on, and correct it where that is not the case.
We also visualize all outlier objects in terms of size and relative location to the ego vehicle to manually correct them (construction vehicles are found to be the most irregular sized objects).
Finally, we check that each object has a complete set of attributes that is consistent with its class and manually correct them where this is not the case.

\subsection{Maps}
\label{sec:dataset_maps}
High-Definition (HD) maps are highly accurate maps that provide a geometric 3D model of the world in centimeter-level precision; and semantics of static road elements such as traffic lights, road markings and road boundaries. 
These HD maps are a critical component for autonomous vehicles, especially in urban environments where precise localization and understanding of road structure are essential.
The nuScenes dataset provides two types of maps: \textit{geometric maps}, represented as 2D bitmaps, and \textit{semantic maps}, which consist of annotated 2D geometric shapes with associated attributes.
This rich information can be leveraged to enhance perception, prediction and planning algorithms. 
In this section, we describe the process of generating geometric and semantic maps, and present the resulting maps along with their potential use cases.

\textbf{Geometric Maps.}
Geometric maps are high accuracy 3D world representation of the environment. 
nuScenes maps are built using a proprietary graph-SLAM engine~\cite{grisetti2010graph-based-slam} that generates globally accurate maps with an accuracy of up to 10 cm.
Mapping is performed using our autonomous vehicles equipped with the same sensor suite used for data collection.
These vehicles accumulate lidar scans and fuse them with GNSS and IMU data to reconstruct the environment as a dense 3D point cloud and use it to localize themselves.
In our graph-SLAM formulation, each node corresponds to a lidar scan, and \textit{factors} represent relative poses between pairs of scans.
During map building, we add two types of factors between point clouds. 
The first is \textit{matching factors}, these are added between consecutive point clouds, rebuilding the AV trajectory. 
The second type is \textit{loop closure factor}~\cite{LoopClosureSurvey}. 
They serve to correct the map from drift, and avoid inconsistencies.
\textit{Loop closure factors} are obtained by associating point clouds corresponding to a same location but captured at different time instances. 
The pose of \textit{loop closure} and \textit{matching factors} are obtained by performing point cloud registration.
We use Iterative Closest Point (ICP)~\cite{ICP} an iterative algorithm that aims at minimizing the metric distance between associated points of the two point clouds.

When our AVs are deployed to build the maps, route planning on the new area is first conducted to optimize the results of the mapping engine as well reduce operational costs. 
A buffer on the start and end point is applied to ensure that all sensors are synchronized. 
A minimum of two passes for both unidirectional roads and bidirectional roads is desired to achieve reliable loop closures. 
Data collection does not take place during and after rain.
Off-peak hours are desired to avoid traffic congestion and large object density.
Finally, we drive with a maximum speed of 40 km/h.

\begin{figure}[ht]
\centering
\includegraphics[width=\linewidth, trim=0 1cm 0 0, clip]{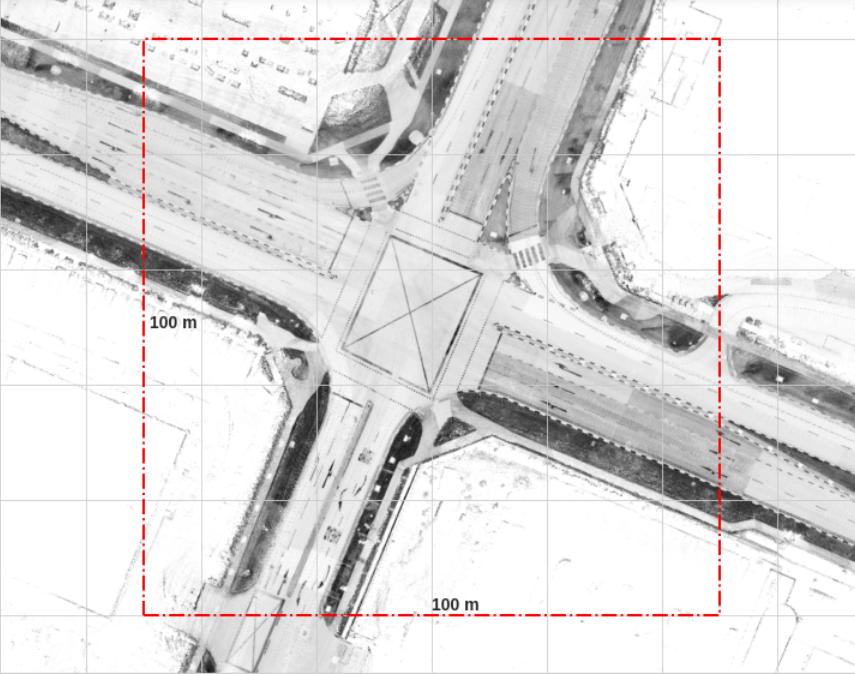}
\caption{Example intensity basemap of an intersection in Singapore One North, obtained from our mapping data collection.}
\label{fig:intensity_map}
\end{figure}

From the resulting 3D maps, we generate 2D top-down representations known as \textit{basemaps}, which are used to assist in semantic map annotation.
Fig.~\ref{fig:intensity_map} shows an example of a basemap represented by lidar intensity values where visual cues such as road boundaries and road markings are clearly visible.

\begin{figure}[ht]
\centering
\includegraphics[width=\linewidth]{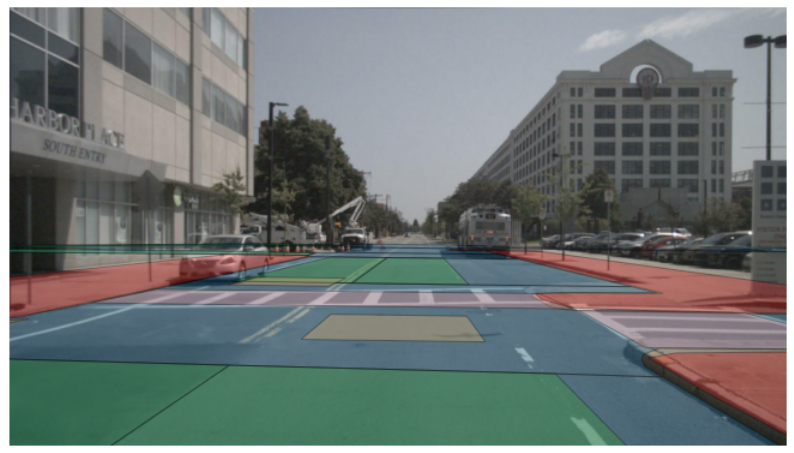}
\caption{Semantic map annotations overlayed on a front camera image from a driving scene in Boston-Seaport.}
\label{fig:map_in_image}
\end{figure}

\textbf{Semantic Maps.} 
Semantic maps provide rich semantic information (e.g. lane elements and traffic lights) on top of the basemaps. 
They are annotated by human experts using an in-house annotation tool with a strict QA process.
Various basemap layers are used as annotation reference.
These present visual cues such as elevation and reflectance, to aid annotators distinguish different road elements and objects.
The precision of these rasters is 10cm/pixel and our semantic map QA only tolerates a margin error of 1 pixel.
Fig.~\ref{fig:map_in_image} shows the semantic map annotations overlayed on the front camera.
We can see that for nearly planar scenes, the alignment between maps and sensor data is accurate.
As shown in the figure, each map element is geometrically represented as points, polylines, or polygons. 

\textbf{Deliverables.}
nuScenes provides maps for 4 locations across 2 cities: Boston Seaport, as well as Singapore OneNorth, Queenstown and Holland Village.
Metadata is provided to convert local coordinates into both geographic (WGS) and projected (UTM) coordinate systems.

The nuScenes devkit includes a \href{https://github.com/nutonomy/nuscenes-devkit/blob/master/python-sdk/tutorials/map\_expansion\_tutorial.ipynb}{Map API} that provides several useful utilities for working with HD maps. 
It allows users to access different map layers based on a specific ego-pose, area, or point of interest. 
The API also offers rendering functions for visualizing map elements in various coordinate frames, making it easier to interpret the spatial context of a scene or sample. 
It can generate rasterized binary masks from selected map layers.
Finally, the API includes tools to support navigation within the road network, such as retrieving lane centerlines and determining connections between lanes and lane connectors.

With this rich map data, there are multiple avenues of research that can and have been explored.
These map data can be additional priors to assist in perception tasks, either as input or as an auxiliary loss in object detection~\cite{hdnet, mp3, Menet, LaneFusion}. 
In addition, the maps also support new tasks such as bird's eye view~(BEV) semantic mapping~\cite{hdmapnet, vectormapnet, bevsegformer, neural_map_planner}, lane detection~\cite{Lane1, Lane2}, and trajectory prediction~\cite{leveraging_trajectory_prediction}.

\subsection{Dataset extensions}
\label{sec:dataset_extensions}
The nuScenes dataset includes two extensions - Panoptic nuScenes~\cite{fong2022nuscenes-panoptic} and nuImages. 

\textbf{Panoptic nuScenes.} 
Panoptic nuScenes~\cite{fong2022nuscenes-panoptic} extends nuScenes by adding point-level annotations to the 1.1B points across the 1000 scenes in nuScenes. The point-level annotations include semantic and instance labels.
Panoptic nuScenes adds 7 \textit{stuff} classes, such as vegetation and drivable surface, on top of the 23 \textit{thing} classes in nuScenes.
This enables the study of interaction of objects (things) with their environment (stuff).
In Panoptic nuScenes, each annotation lidar point is manually annotated with a semantic label. 
To reduce the annotation effort, the 3D boxes from nuScenes are used to initialize the semantic labels of the points for objects belonging to the thing classes. 
Points that belong to stuff classes are manually annotated.
Thereafter, refinement is manually done for points which are included in multiple thing boxes, or points which are close to stuff points (e.g. vehicle wheels close to the ground plane). 
The point-level labels are then combined with the 3D bounding boxes from nuScenes to obtain instance labels for each point. An instance consists of the points that fall within a 3D bounding box and have the same segmentation class as the box. 
For bounding boxes that overlap, the points in the overlaps are labeled as \emph{noise}.  
The percentage of points that are present in such overlapping regions is less than 0.8\% for all classes. 
Using the instance ID of the boxes, the instances are ensured to be temporally consistent. 
This approach eliminates a significant number of points which need to be manually annotated. 
Like with nuScenes, multiple rounds of validation are performed to achieve high-quality segmentation annotations.
For more details, we refer the reader to~\cite{fong2022nuscenes-panoptic,caesar2016cocostuff}.

\textbf{nuImages.} 
Following the publication of nuScenes, many users expressed interest in working on the more traditional 2D detection and segmentation, rather than the 3D detection in nuScenes. 
To address this, we released the nuImages dataset.
nuImages is a standalone dataset with annotated 93,000 images, all of which are annotated with instance masks, 2D bounding boxes and semantic segmentation masks.
Annotating the 1.4 million images in nuScenes, or a subset thereof, with 2D annotations would have led to highly redundant annotations, which are not ideal for training an object detector. 
We thus labeled a more varied image dataset from nearly 500 logs (compared to 83 in nuScenes) of driving data.
nuImages uses the same license as nuScenes.

The 93,000 images are selected using two different approaches. 
We use active learning techniques to select about 75\% of the images to be challenging according to the uncertainty of an image-based object detector, with a particular focus on 
1) rare classes like bicycles, 
2) adverse conditions like rain, snow and night-time and 
3) a balanced spatio-temporal distribution.
The remaining 25\% of the images are sampled uniformly to guarantee a representative dataset and avoid strong biases. 
After careful review some images are discarded due to camera artifacts, as they are too light or too dark.
Furthermore, to preserve the privacy of pedestrians in the dataset, we manually review the 1k images with the largest pedestrian bounding boxes.
We then remove around 50 camera images that are privacy sensitive, due to the pedestrians being close to the camera.

While lidar data was originally meant to be part of the dataset, changing procedures for calibration and synchronization over the data collection period made it challenging to guarantee a good alignment between the lidar and camera data.
Additionally we include 6 past and 6 future camera images at 2 Hz for each annotated image to study the temporal dynamics. 
This means that nuImages contains 93k video clips with 13 frames spaced out at 2 Hz resulting in a total of 1.2M camera images.
To date, this scale and diversity is only matched by the BDD100K~\cite{Bdd100k} dataset with 100k videos.

\subsection{Leaderboards}
\label{sec:leaderboards}
This section describes design choices behind the leaderboards in nuScenes. 
While details are specific to the detection leaderboard, most insights transfer to the leaderboards of other tasks as well.

The leaderboard of the KITTI~\cite{geiger2012kitti} dataset consists of different strata. Performance is listed separately per class.
Depending on a combination of box height, occlusion and truncation level, objects are categorized by their difficulty as easy, medium or hard.
The performance for each class is listed in a separate leaderboard.
While these strata are useful for evaluation, the lack of an aggregate metric over all classes or difficulties, leads to most works in the literature just comparing a single class and difficulty (car, medium), thus effectively ignoring a big part of the dataset.
This is a problem as no overall measure of the AV performance is provided.
Furthermore, some users may selectively pick the metric that favors their method.
Finally, since the focus is usually on the car class, rare classes are neglected. 
On the other hand, the leaderboard of the Waymo~\cite{sun2019wod} dataset allows users to specify an arbitrary number of past and future frames that are used by the method, thus lacking standardization and making results incomparable.
We created a single leaderboard for each task to make all submissions directly comparable.
While strata (class, distance, confidence etc.) do exist for more detailed evaluation, we do not use them in the main metric or on the default leaderboard view to reduce complexity.

We further provide a search function for methods and their release date and descriptions, as well as filters to select the sensor modalities (the top three combinations are lidar-only, camera-only, camera-lidar), use of map data (0\% of all submissions) and use of external data (20\% of all submissions).
%
Finally, while metrics on some leaderboards are unintuitive and in no clearly defined relationship to each other, metrics on the nuScenes leaderboard are decomposable on multiple levels.
Our main metric, the nuScenes Detection Score (NDS)~\cite{caesar2020nuscenes}, is a linear combination of the well-established mAP and five true-positive metrics, thus easily decomposable into its submetrics.
Each submetric in this linear combination is an average over all classes, thus the contribution of each class and metric is immediately obvious.
\section{Related Datasets}
\label{sec:relateddatasets}
In this section we discuss datasets released after the original nuScenes release. We analyze which aspects of these datasets were inspired by nuScenes and how they further improved upon it. 
For an overview of prior datasets, we refer the reader to the original nuScenes paper~\cite{caesar2020nuscenes}. 
 
\subsection{Multimodal Perception}
Table~\ref{table:multimodal_perception} provides an overview of current multimodal datasets and the tasks they support. 
Most multimodal datasets equip vehicles with a combination of lidar and camera sensors. 
These datasets annotate 3D bounding boxes over time and therefore support the tasks of 3D object detection and tracking, arguably the main tasks in automotive perception.
Explicit 2D box annotations are less common, as 3D boxes can be projected onto image planes, though these projections rarely produce tightly fitting 2D boxes, limiting their usefulness for dedicated 2D detection tasks.

Focusing on bounding boxes provides detailed information about foreground objects, but the surrounding environment is not taken into account.
Capturing this context requires pixel- or point-level annotations, enabling instance or panoptic segmentation.
Covering both foreground and background elements, such annotations are considerably more labor-intensive than bounding boxes. 
As a result, large-scale multimodal datasets with dense segmentation labels remain scarce~\cite{caesar2020nuscenes,semantic_kitti,sun2019wod}.

Multimodal datasets in autonomous driving often target specific tasks.
For instance, Lyft Level 5 \cite{Lyft} focuses on motion planning, while ONCE \cite{ONCE} provides large-scale unlabeled data for self-supervised learning. 
In contrast, nuScenes~\cite{caesar2020nuscenes} was designed as a comprehensive multimodal dataset covering multiple tasks from the start, including 3D detection, tracking, prediction, lidar and panoptic segmentation across cameras, LiDARs and radar sensors.
Other large-scale datasets provide a subset of these tasks.
ApolloScape \cite{Apolloscape} offers high-resolution 3D data for detection and segmentation but does not provide point cloud segmentation labels necessary for full 3D semantic segmentation.
Waymo Open Dataset (WOD) \cite{sun2019wod} is comparable to nuScenes in terms of the number of scenes and tasks, but it lacks radar data.
KITTI-360 \cite{KITTI_360} and AIODrive \cite{AIODrive} include multimodal sensor data but do not provide mapping annotations, limiting their use for mapping-based tasks.

\begin{sidewaystable*}
\centering \footnotesize
\caption{
Comparison of supported tasks/benchmarks in multimodal datasets. 
To support a task, a dataset needs to provide necessary annotations and any benchmarks for it.
*C/L/R: Camera/Lidar/Radar; Occ: Occlusion; Topol: Topological; VQA: Visual Question Answering
}
\begin{tabular}{| >{\centering\arraybackslash}m{2.2cm} | c | c | >{\centering\arraybackslash}m{3.5cm} | c | c | c | c | c | c | c | c | c |}
\hline
\multirow{2}*{Dataset} & \multirow{2}*{Year} & Sensors & Main & Det & Seg & Track & Pan Seg & Map & Pred/Plan/ & \# Bench- \\ 
& & C/L/R & Highlights & 2D/3D & 2D/3D & 2D/3D & 2D/3D/4D & Lane/Semantic/Occ/Top & VQA & marks \\
\hline
  KITTI \cite{geiger2012kitti} & 2012 & \cmark/\cmark/\xmark & First multimodal AV dataset & \cmark/\cmark & \cmark/\cmark\cite{semantic_kitti} & \cmark/\cmark & \xmark/\cmark\cite{semantic_kitti}/\cmark\cite{semantic_kitti} & \cmark/\xmark/\xmark/\xmark & \cmark/\xmark/\cmark\cite{wu2023referringmultiobjecttracking} & 17 \\  
 \hline
 KAIST \cite{choi2018kaist} & 2018 & \cmark/\cmark/\xmark & Day/Night 2D Detection & \cmark/\xmark & \xmark/\xmark & \xmark/\xmark & \xmark/\xmark/\xmark & \xmark/\xmark/\xmark/\xmark & \xmark/\xmark/\xmark & 0 \\
 \hline
 ApolloScape \cite{Apolloscape} & 2018 & \cmark/\cmark/\xmark & Support Inpainting & \xmark/\cmark & \cmark/\xmark & \xmark/\cmark & \xmark/\xmark/\xmark & \cmark/\cmark/\xmark/\xmark & \cmark\cite{ma2019trafficpredict}/\xmark/\xmark & 8 \\ 
 \hline
 Honda 3D \cite{Honda3D} & 2019 & \cmark/\cmark/\xmark & 3D Perception & \xmark/\cmark & \xmark/\xmark & \xmark/\cmark & \xmark/\xmark/\xmark & \xmark/\xmark/\xmark/\xmark & \xmark/\xmark/\xmark & 0 \\
 \hline
 WoodScape \cite{yogamani2019woodscape} & 2019 & \cmark/\cmark/\xmark & Multi-Fisheye Cameras & \cmark/\cmark & \cmark/\xmark & \xmark/\cmark & \xmark/\xmark/\xmark & \xmark/\xmark/\xmark/\xmark & \xmark/\xmark/\xmark & 0 \\
 \hline
 \rowcolor{green!15}
 \textbf{nuScenes} \cite{caesar2020nuscenes} & 2019 & \cmark/\cmark/\cmark & Radar, $360^{\circ}$ Coverage, Semantic maps & \cmark/\cmark & \cmark/\cmark & \cmark/\cmark & \cmark/\cmark/\cmark & \cmark/\cmark/\cmark\cite{tian2023occ3d}/\cmark\cite{Openlanev2} & \cmark/\xmark \cite{karnchanachari24nuplan}/\cmark\cite{Talk2Car} & 11 \\
 \hline
 Argoverse \cite{argoverse1} & 2019 & \cmark/\cmark/\xmark & Detailed Semantic \& HD Maps & \xmark/\cmark & \xmark/\xmark & \xmark/\cmark & \xmark/\xmark/\xmark & \cmark/\cmark/\xmark/\xmark & \cmark/\xmark/\xmark & 4 \\
 \hline
 Lyft L5 \cite{Lyft} & 2019 & \cmark/\cmark/\xmark & Prediction \& Planning & \xmark/\cmark & \xmark/\xmark & \xmark/\cmark & \xmark/\xmark/\xmark & \cmark/\cmark/\xmark/\cmark & \cmark/\cmark/\xmark & 2 \\
 \hline
  WOD \cite{sun2019wod} & 2019 & \cmark/\cmark/\xmark & Diverse Geography & \cmark/\cmark & \cmark/\cmark & \cmark/\cmark & \cmark/\xmark/\xmark & \cmark\cite{OpenLane}/\cmark/\cmark\cite{tian2023occ3d}/\cmark & \cmark\cite{Kan_2024_icra}/\cmark\cite{Kan_2024_icra}/\cmark\cite{reason2drive} & 21 \\  
 \hline
  A*3D \cite{A*3D} & 2019 & \cmark/\cmark/\xmark & Diverse Time \& Weather & \xmark/\cmark & \xmark/\xmark & \xmark/\xmark & \xmark/\xmark/\xmark & \xmark/\xmark/\xmark/\xmark & \xmark/\xmark/\xmark & 0 \\  
 \hline
  A2D2 \cite{a2d2} & 2019 & \cmark/\cmark/\xmark & 2D/3D Segmentation & \cmark/\cmark & \cmark/\cmark & \xmark/\xmark & \xmark/\xmark/\xmark & \xmark/\xmark/\xmark/\xmark & \xmark/\xmark/\xmark & 0 \\
 \hline
  ONCE \cite{ONCE} & 2019 & \cmark/\cmark/\xmark & Unsupervised Detection & \xmark/\cmark & \xmark/\xmark & \xmark/\xmark & \xmark/\xmark/\xmark & \cmark\cite{yan2022once}/\xmark/\xmark/\xmark & \xmark/\xmark/\cmark\cite{reason2drive} & 2 \\
 \hline
  Cirrus \cite{cirrus} & 2019 & \cmark/\cmark/\xmark & Long-range Detection & \xmark/\cmark & \xmark/\xmark & \xmark/\xmark & \xmark/\xmark/\xmark & \xmark/\xmark/\xmark/\xmark & \xmark/\xmark/\xmark & 0 \\
 \hline
  AIODrive \cite{AIODrive} & 2020 & \cmark/\cmark/\cmark & Synthetic Dataset & \cmark/\cmark & \cmark/\cmark & \cmark/\cmark & \xmark/\xmark/\xmark & \cmark/\cmark/\xmark/\xmark & \cmark/\xmark/\xmark & 0 \\
 \hline
  CADC \cite{CADC} & 2020 & \cmark/\cmark/\xmark & Adverse Weather & \xmark/\cmark & \xmark/\xmark & \xmark/\xmark & \xmark/\xmark/\xmark & \xmark/\xmark/\xmark/\xmark & \xmark/\xmark/\xmark & 0 \\
 \hline
  nuPlan \cite{karnchanachari24nuplan} & 2021 & \cmark/\cmark/\xmark & Prediction \& Planning & \xmark/\cmark & \xmark/\xmark & \xmark/\cmark & \xmark/\xmark/\xmark & \cmark/\cmark/\xmark/\cmark & \cmark/\cmark/\xmark & 3 \\
 \hline
  PandaSet \cite{pandaSet} & 2021 & \cmark/\cmark/\xmark & Forward-facing Lidar & \xmark/\cmark & \xmark/\cmark & \xmark/\xmark & \xmark/\xmark/\xmark & \xmark/\xmark/\xmark/\xmark & \xmark/\xmark\xmark & 0 \\
 \hline
  Argoverse 2 \cite{argoverse2} & 2021 & \cmark/\cmark/\xmark & Include Map Change & \xmark/\cmark & \xmark/\xmark & \xmark/\cmark & \xmark/\xmark/\xmark & \cmark/\cmark/\cmark/\cite{khurana2023point}/\cmark\cite{Openlanev2} & \cmark/\xmark/\xmark & 6 \\
 \hline
  VOD \cite{vod} & 2022 & \cmark/\cmark/\cmark & 4D Radar & \xmark/\cmark & \xmark/\xmark & \xmark/\cmark & \xmark/\xmark/\xmark & \cmark\cite{VOD-prediction}/\cmark\cite{VOD-prediction}/\cmark/\xmark & \cmark\cite{VOD-prediction}/\xmark/\xmark & 0 \\
 \hline
  K-Radar \cite{KRadar} & 2022 & \cmark/\cmark/\cmark & 4D Radar & \xmark/\cmark & \xmark/\xmark & \xmark/\xmark & \xmark/\xmark/\xmark & \xmark/\xmark/\xmark/\xmark & \xmark/\xmark/\xmark & 0 \\ 
 \hline
  KITTI-360 \cite{KITTI_360} & 2022 & \cmark/\cmark/\xmark & 3D Perception & \xmark/\cmark & \cmark/\cmark & \xmark/\xmark & \xmark/\xmark/\xmark & \xmark/\xmark/\xmark/\xmark & \xmark/\xmark/\xmark & 12 \\    
 \hline
  aiMotive \cite{aimotive} & 2022 & \cmark/\cmark/\cmark & Long-range Detection & \xmark/\cmark & \xmark/\xmark & \xmark/\xmark & \xmark/\xmark/\xmark & \xmark/\xmark/\xmark/\xmark & \xmark/\xmark/\xmark & 0 \\
  \hline
  OpenMPD \cite{OpenMPD} & 2022 & \cmark/\cmark/\xmark & 2D \& 3D Perception & \cmark/\cmark & \cmark/\xmark & \xmark/\xmark & \xmark/\xmark/\xmark & \cmark/\xmark/\xmark/\xmark & \xmark/\xmark/\xmark & 3 \\ 
 \hline
  Ithaca365 \cite{Ithaca365} & 2022 & \cmark/\cmark/\xmark & Adverse Weather & \xmark/\cmark & \cmark/\xmark & \xmark/\xmark & \xmark/\xmark/\xmark & \cmark/\xmark/\xmark/\xmark & \xmark/\xmark/\xmark & 0 \\
 \hline
  Boreas \cite{Boreas} & 2023 & \cmark/\cmark/\cmark & Adverse Weather & \xmark/\cmark & \xmark/\xmark & \xmark/\xmark & \xmark/\xmark/\xmark & \xmark/\xmark/\xmark/\xmark & \xmark/\xmark/\xmark & 6\\ 
 \hline
  ZOD \cite{zenseact} & 2023 & \cmark/\cmark/\cmark & Diverse Geography \& Long-range & \cmark/\cmark & \cmark/\xmark & \xmark/\xmark & \xmark/\xmark/\xmark & \cmark/\xmark/\xmark/\xmark & \xmark/\xmark/\xmark & 0 \\
 \hline
  BSD~\cite{boschstreet} & 2024 & \cmark/\cmark/\cmark & 4D Radar Resolution \& Scale & \xmark/\cmark & \xmark/\xmark & \xmark/\xmark & \xmark/\xmark/\xmark & \xmark/\xmark/\xmark/\xmark & \xmark/\xmark/\xmark & 0 \\
 \hline
  TruckScenes \cite{fent2024man} & 2024 & \cmark/\cmark/\cmark & 4D Radar \& Autonomous Trucks & \xmark/\cmark & \xmark/\xmark & \xmark/\xmark & \xmark/\xmark/\xmark & \xmark/\xmark/\xmark/\xmark & \xmark/\xmark/\xmark & 0 \\
\hline
  AevaScenes \cite{aevascenes} & 2025 & \cmark/\cmark/\xmark & FMCW LiDAR & \xmark/\cmark & \xmark/\cmark & \xmark/\xmark & \xmark/\xmark/\xmark & \cmark/\xmark/\xmark/\xmark & \xmark/\xmark/\xmark & 0 \\
 \hline
\end{tabular}
\label{table:multimodal_perception}
\end{sidewaystable*}

\subsection{Radar}
The nuScenes dataset was the first to include automotive radar as part of a multi-modal sensor suite. It features five radars arranged to provide full 360° coverage, each scanning in a horizontal plane. Radar sensors offer a low-cost and weather-robust alternative to lidar, especially in Advanced Driver Assistance Systems (ADAS).
%
Despite these advantages, radar data in nuScenes presents several challenges.
The radar sensors are not synchronously triggered with the other sensors, resulting in significant temporal misalignments. 
Furthermore, the positional accuracy and resolution are significantly lower than that of the lidar, which makes it very challenging to annotate objects from radar data alone.
Multi-path propagation can lead to spurious returns on wet roads or in windows, thus often confusing non-expert annotators.
At long distances, where no lidar data is available, we found it beneficial to annotate the lateral position and size from the camera. For the longitudinal position we use radar, while for the often unobserved longitudinal size we use prior knowledge about typical sizes of specific vehicle types.
%
A key advantage of radar is its ability to directly estimate the radial component of an object's velocity in the radial direction using the doppler effect.
This leads to lower perception system latency by reducing the need for temporal aggregation, which is usually done for camera or lidar.
We observe that in some cases the doppler vectors do not always point to the radar sensor in nuScenes, which may be due to minor calibration and synchronization errors, as well as the radar's ego motion compensation using unfiltered odometry data, which differs from the provided lidar-based localization data.

Since the release of nuScenes (2019), numerous other datasets have also included radar.
Astyx HiRes2019 (2019) was the first dataset to use a (proprietary) 4D radar, thus addressing an important shortcoming of the 3D radar in nuScenes which cannot capture elevation information. However, the dataset became unavailable following the company’s acquisition.
The View of Delft~\cite{delft} dataset (2022) was then made available to the public, featuring 4D radar and challenging interactions with Vulnerable Road Users in the Netherlands.
Similarly, TJ4DRadSet~\cite{tj4dradset} captures dense driving sense in a city in China using 4D radar data.
K-Radar~\cite{KRadar} (2022) goes beyond nuScenes by providing raw 4D radar cube data (range-doppler-azimuth), enabling low-level radar learning, and included a broader range of road types while adopting nuScenes-like 3D object annotations.
The Bosch Street Dataset~\cite{boschstreet} (2024) is more than 100x larger than existing 4D radar datasets, but is not available to the general public.
TruckScenes~\cite{truckscenes2024} (2024) is the first dataset for autonomous trucking applications, featuring 4D radars at elevated viewpoints and unique challenges due to occlusions from the truck's trailer.
Several datasets rely on novel types of radars.
The Oxford Radar RobotCar dataset~\cite{RadarRobotCarDatasetICRA2020} (2019) uses a Navtech CTS350-X scanning radar and is well suited for radar-based odometry. Due to its revolving nature can capture data from all directions at high resolution using only a single sensor. However, the data is captured at a relatively low frequency of 4 Hz and the dataset lacks 3D annotations, thus making it less suitable for perception. 
The Zendar dataset~\cite{zendardataset} (2020) increases radar resolution by using Synthetic Aperture Radar (SAR) with a proprietary radar system. However, this technology only works for slow moving objects due to the long integration times required by SAR.
Collectively, these datasets address the limitations of nuScenes' radar setup, while keeping many of its conventions around annotations, data formats, devkit and vehicle setup.

\subsection{Mapping}
Detecting road elements and their topology is important to have full geometric and semantic understanding of the environment. Hence, a plethora of mapping benchmark datasets have been introduced prior and following the introduction of the nuScenes dataset.
Datasets can range from simple road elements that could aid in self-localization to a complex list of road element types and attributes to aid in path planning and reasoning.
Earlier datasets follow the former where it primarily focuses on lane detection~\cite{Apolloscape,Bdd100k,culane}. However, these are limited to 2D annotations in camera view that need to be translated to BEV or 3D view in order to be fully useful in downstream modules or be leveraged in end-to-end perception models. 
And while some datasets extended these works to have 3D lane annotations~\cite{OpenLane, Once3D, GenLanenet}, these works are either synthetically or automatically generated by some heuristics which limits their accuracy. 
Moreover, prior datasets are limited to the image-modality which puts a cap on the potential of the benchmark compared to multi-modal sensors. 
Hence, large-scale multi-modal AV datasets with diverse and accurate semantic maps are desired to further advance the mapping field.

nuScenes is one of the few large-scale datasets that provided rich map data alongside Lyft~\cite{Lyft}, Argoverse v2~\cite{argoverse2} and WOD~\cite{sun2019wod}. These datasets provide semantic maps that are manually annotated by human experts on top of the localized sensor map. 
 As presented in Section~\ref{sec:dataset_maps}, nuScenes contains essential vectorized map layers such as drivable area, lanes \& dividers, lane connectivities and pedestrian crossings; which is similar to the two mentioned datasets.
 Differently, nuScenes has other relevant layers such as traffic lights and its lane associations, walkways, stop lines, and carpark area.
 Lyft on the other hand also has extensive map data that extends to road classes, road paintings, speed limits, lane restrictions, traffic signs, restrictions and speed bumps. While Argoverse v2 has limited but concise map layers in 3D lane geometry, paint markings, and crosswalks. Aside from semantic maps, Argoverse v2 includes real-valued ground height at thirty centimeter resolution to aid in ground height filtering or detecting road elements in 2.5D. In addition, they also provide a map change detection dataset~\cite{MapChangeDetection} to support research on map changes caused by mismatch between online map detections and HD maps due to environmental changes or human annotator error. 
 Finally, WOD do not provide full HD maps of a region but provides map data for each scenario. This include dynamic map states (e.g. traffic signal states and corresponding lane ids) and map features (lane centers, lane boundaries, road boundaries, crosswalks, speed bumps, and stop signs).

Some works also extend nuScenes further to provide an avenue for more research. For example~\cite{Lanelet2fornuScenes} converted nuScenes vector layers to Lanelet2 primitives~\cite{Lanelet2} which makes it more flexible for prediction and planning research. In addition, it enhances the semantic maps by providing spatial semantic information as well as diverse map-based anchor paths. 
OpenLane v2~\cite{Openlanev2} builds on top of both nuScenes and Argoverse and adds new annotations for road elements and its corresponding topology relationships. In particular, this dataset provides manually annotated and validated traffic elements such as road signs, road painting, and traffic lights with real-time status as well as its association to the centerlines to represent traffic rules. This introduces a new task of topology understanding from multi-view images. 

\subsection{Semantic Occupancy}
Semantic Occupancy tasks typically involve dividing a driving scene into a 3D grid and predicting a semantic label for each voxel, either for a present or future timestamp.
Occ3D-nuScenes~\cite{tian2023occ3d} and OpenOcc~\cite{sima2023occnet} build 3D occupancy benchmarks on top of nuScenes and Panoptic nuScenes. Both utilize similar automatic labeling pipelines to derive the semantic occupancy ground-truth for each voxel in each scene. Both pipelines consists of these components: a) accumulation of lidar sweeps to increase the density of the lidar point cloud, b) ray-casting to determine if each voxel can be seen by each camera on the ego, to determine occupied and free voxels and c) refinement of the labels by completing the scene (i.e. filling in gaps in the occupancy map).

\subsection{Language}
There has recently been much interest in incorporating language into visual tasks, including autonomous driving.
Talk2Car \cite{deruyttere2019talk2car} was the first to expand nuScenes \cite{caesar2020nuscenes} with natural language commands based on a reference object found in a scene. 
It attempts to simulate the setting in which a passenger in the autonomous vehicle issues verbal commands to control the vehicle.
nuScenes-QA \cite{qian2023nuscenesqa} creates a visual question answering (VQA) task specifically for autonomous driving based on nuScenes \cite{caesar2020nuscenes}.
VQA for autonomous driving makes for a challenging task as the raw visual data is multi-modal and encapsulates agent dynamics from multiple frames.
nuScenes-QA \cite{qian2023nuscenesqa} focuses on scene-level comprehension through language.
nuPrompt \cite{wu2023nuprompt} provides a more fine-grained semantic understanding of a scene by using a language prompt to describe groups of objects, which are similar in some manner. 
This correspondence between the 3D instances and text builds upon the 3D, multi-view and multi-frame nature of nuScenes \cite{caesar2020nuscenes}.
More recently, DriveLM~\cite{drivelm} add annotations to nuScenes to integrate Graph VQA into end-to-end planning and answer questions about objects in the environment and how to navigate around them.
nuScenes-MQA~\cite{nuscenesmqa} use markup annotations to learn general sentence concepts rather than just sequences of words.
OmniDrive~\cite{omnidrive} uses counterfactual reasoning to provide denser supervision signals that bridge planning trajectories and language-based reasoning.
To address a gap of existing Vision-Language Models~(VLMs), nuScenes-SpatialQA~\cite{nuscenesspatialqa} create a new benchmark that highlights spatial understanding of VLMs.

\section{Tasks and Metrics}
\label{sec:tasksandmetrics}

In this section we describe the main tasks of the nuScenes dataset, as well as their metrics. 

\subsection{Detection}
3D object detection is the task of localizing objects from a set of predefined classes in space and classifying them accordingly.
Numerous metrics are used for this task in the literature.
Average Precision (AP), with a threshold on intersection-over-union (IOU), is commonly used to measure object detection performance \cite{geiger2012kitti, argoverse2}. 
However, this metric entangles location, size and orientation estimates. 
ApolloScape \cite{huang2018apolloscape}, in its 3D car instance challenge, tries to decouple these by defining thresholds for each error type and recall threshold, but this results in 10 x 3 thresholds which makes such an approach complex, arbitrary and unintuitive.

For the nuScenes detection task, the primary metric used to measure detection performance is the nuScenes Detection Score (NDS) \cite{caesar2020nuscenes}.
NDS includes, in addition to mean AP (mAP), a set of five mean True Positive metrics - Average Translation Error (ATE), Average Scale Error (ASE), Average Orientation Error (AOE), Average Velocity Error (AVE), and Average Attribute Error (AAE). 
NDS is thus able to measure detection performance, while also taking into account the quality of detections in terms of box location, size, orientation, attributes and velocity.
It has been found that NDS has a higher correlation to the downstream planning performance than mAP \cite{schreier2023offline}.

Following an update in 2020, the nuScenes detection task now also measures Planning KL-Divergence (PKL) \cite{pklmetric,guo2020nuscenes-pkl}. 
This metric measures detection performance by measuring differences between how a planner would plan when given detections from a detector instead of human-labeled detections. A PKL of 0 corresponds to an optimal detector. 
A larger PKL score corresponds to worse detection performance. 

More recently, WOD \cite{sun2019wod} introduced the APH metric, which incorporates heading into the AP metric. However, APH does not account for other aspects of detection which are critical in autonomous driving, such as velocity and attribute estimates.
Follow-up work by \cite{hung2024} introduces Longitudinal Error Tolerant 3D Average Precision (LET3D-AP), a metric that is more suitable for evaluation of monocular 3D object detectors, since it is more forgiving to depth estimation errors. Using this metric, they show that monocular 3D object detectors can even outperform lidar-based detectors.

\subsection{Tracking}
3D object tracking is the task of continuously estimating the position, class and identity of every object in the scene from a set of predefined classes.
In the nuScenes tracking task, Average Multi Object Tracking Accuracy (AMOTA) \cite{Weng2020AB3DMOT} is the primary metric by which performance is ranked. 
AMOTA is able to evaluate a tracker across a range of scenarios by averaging the MOTA scores across different recall values. In contrast, most other metrics only evaluate the tracker's performance at the optimal MOTA threshold. However, there are no metrics that explicitly cover the maximum recall, which is essential for understanding the tracker's performance in situations which call for a very low number of False Negatives.

Besides AMOTA, two novel metrics are also used to measure tracking performance: track initialization duration (TID) and longest gap duration (LGD). TID measures the duration from the beginning of the track until the time an object is first detected, since a number of trackers require a fixed window of past sensor readings to get a good initialization. LGD computes the longest duration of any detection gap in a track, and is highly relevant for autonomous driving, where missing an object for several seconds is less acceptable than having many fragmented short-term tracks.

More recently, the Higher Order Tracking Accuracy (HOTA) \cite{luiten2020hota} metric was developed to provide a balanced evaluation of multi-object tracking performance by jointly measuring detection, association, and localization accuracy. This addresses the limitations of metrics like AMOTA and AMOTP by providing a unified score that reflects overall tracking performance.

\subsection{Lidar (Panoptic) Segmentation}
Lidar segmentation is the task of predicting the category of every lidar point in a set of point clouds. The annotated Panoptic nuScenes dataset provides up to 32 classes, but a high-level label set of 16 classes (10 foreground and 6 background classes) is used for the official evaluation. It has 28130 samples for training, 6019 for validation and  6008 for testing. 

nuScenes offers two lidar segmentation tasks - lidar semantic segmentation and lidar panoptic segmentation. With lidar semantic segmentation, the class of each point is predicted. IOU is used to evaluate the panoptic segmentation performance, and methods are ranked based on the average IOU (mIOU) over all the 16 classes. Lidar panoptic segmentation goes a step further than lidar semantic segmentation by also requiring the instance of each point to be predicted, in addition to its class. Panoptic Quality (PQ) \cite{kirillov2019panoptic} is used as the primary metric for evaluation.

\subsection{Lidar Panoptic Tracking} 
This task consists of both segmentation and tracking. For panoptic segmentation, the goal is to predict the semantic categories of every point, and additional instance IDs for things. Specifically, for the \textit{Lidar-track}, only current frame and preceding frames within 0.5s are allowed, maximally 10 frames (inclusive of current frame). Moreover, only lidar panoptic annotations from panoptic nuScenes are allowed. For the \textit{Open-track}, all past data is allowed but not future data. All nuScenes, Panoptic nuScenes and nuImages annotations can also be used. 

While panoptic segmentation focuses on static frames, panoptic tracking additionally enforces temporal coherence and pixel-level associations over time. For both tracks, there are 16 categories (10 thing and 6 stuff classes).

\cite{fong2022nuscenes-panoptic} introduces a Panoptic Tracking metric (PAT) to simultaneously assess the performance of both panoptic segmentation and instance association across time. PAT, as a metric, has error-type differentiability, penalizes fragmentation of tracking, has long-term track consistency, penalizes ID transfer, and penalize the removal of correctly segmented instances with incorrect IDs.
\section{Methods}
\label{sec:methods}

As a large-scale full sensor-suite dataset for autonomous driving nuScenes has become the standard benchmark for many AV tasks. 
In particular, the nuScenes homepage officially hosts the following challenges: 3D Detection\footnote{\url{https://www.nuscenes.org/object-detection}}, Tracking\footnote{\url{https://www.nuscenes.org/tracking}}, Lidar Segmentation\footnote{\url{https://www.nuscenes.org/lidar-segmentation}}, Panoptic Segmentation\footnote{\url{https://www.nuscenes.org/panoptic}}.
nuScenes is also used by the community in various other benchmarks, such as 2D image detection via nuImages, Occupancy Prediction and Map Element Detection. 
In this section, we present in more detail the different state-of-the-art methods for these tasks and provide interesting insights on the progress of AV technology.  

\subsection{3D Object Detection}
\begin{figure}[ht]
\centering
\includegraphics[width=\linewidth]{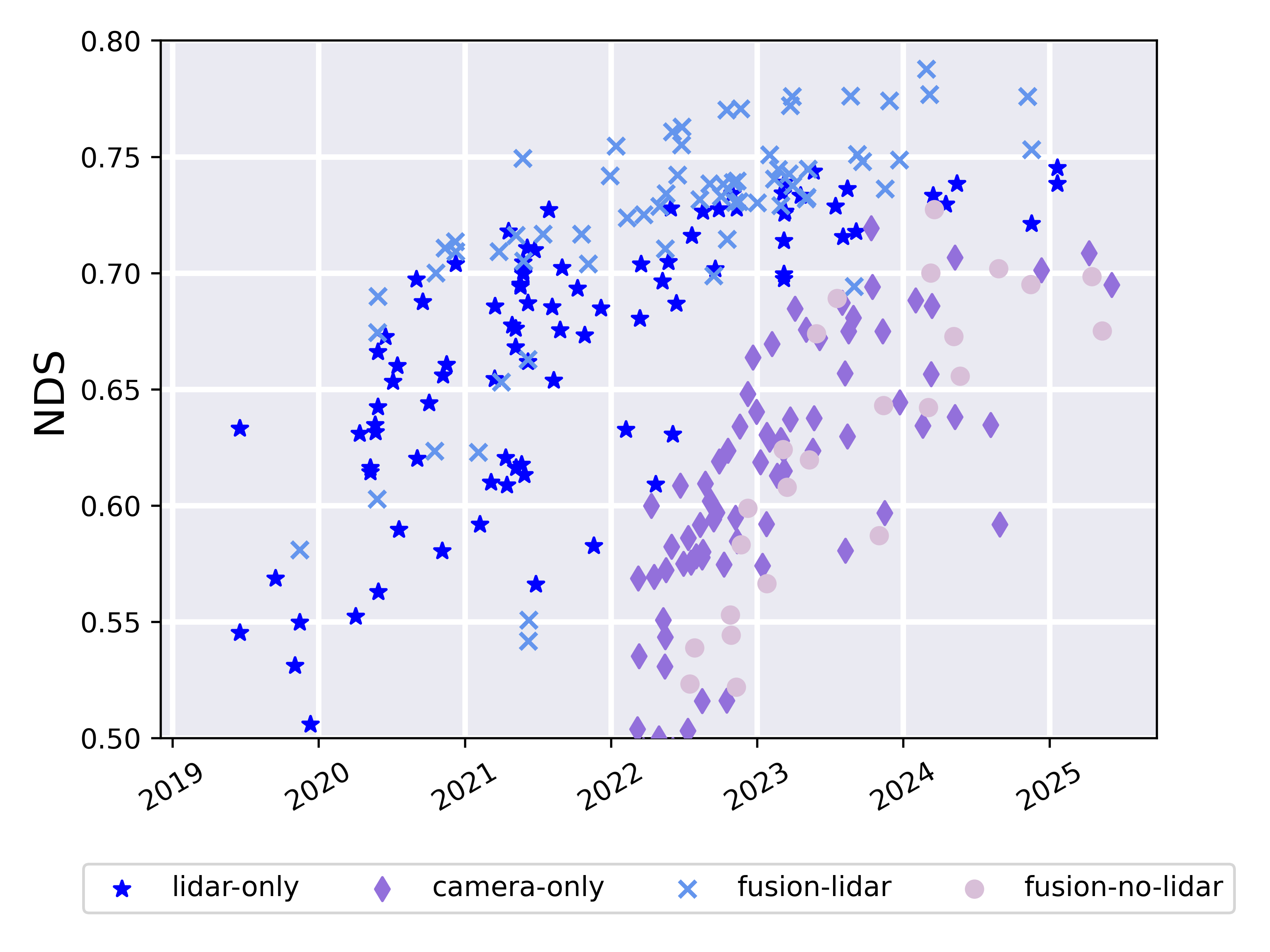}
\vspace{-3mm}
\caption{Results on the nuScenes detection leaderboard.}
\label{fig:challenge_detection}
\end{figure}

Fig.~\ref{fig:challenge_detection} shows the performance of leading methods on the 3D object detection challenge.
We observe a huge improvement in NDS since the challenge began in 2019, which are increasingly using transformers.
More recently, performance improvements are slowing down.
In addition, we observe some interesting trends within the nuScenes community.

\textbf{Camera-radar fusion.}
Research into camera-radar fusion on nuScenes has been gaining momentum. There were no camera-radar fusion methods submitted in the first year of the nuScenes detection challenge. 
The increasing interest in camera-radar fusion in the following years led to a marked improvement in the NDS of such methods.
Notably, the performance of fusion methods, in particular camera-radar fusion which do not make use of lidar at all is catching up to that of lidar-only methods. 
In 2020 the gap between the \emph{lidar-only} method CenterPoint~\cite{yin2021center} and the \emph{camera-radar fusion} method CenterFusion~\cite{nabati2020centerfusion} amounted to $22.4\%$ NDS.
With the advent of HVDetFusion \cite{lei2023hvdetfusion}, a leading camera-radar method that uses uses test-time augmentation (TTA), the gap was only $8.1\%$ NDS, which is an almost 3x reduction in three years.
Due to the remaining performance gap between methods that do use lidar and those that do not, camera-radar fusion is still a promising research direction, considering that lidar sensors are significantly more expensive.

\textbf{Offline detection.}
Manual annotation of 3D bounding boxes is costly and time-consuming.
Therefore more and more works develop perception systems to autolabel datasets.
This is commonly referred to as offline perception~\cite{qi2021offboardperception,karnchanachari24nuplan}.
Such methods use compute intensive algorithms that are not limited to the hardware or real-time constraints of AVs. 
Furthermore, by incorporating future sensor data, additional information can be used during training and inference.
BEVFormer v2 \cite{Yang2022BEVFormerV2} is a camera-based method that employs a bidirectional temporal encoder that uses BEV features from both the past and the future. The use of future information gives a significant improvement in the NDS of the model from 0.498 to 0.529. Sparse4D \cite{lin2022sparse4d} proposes to sparsely sample queries by adding information from past frames to make it 4D sampling as an alternative to dense queries. 
By leveraging four previous frames, it improves the NDS from 0.451 to 0.541. However, its computational complexity increases linearly with the number of past frames. 
To overcome this limitation, Sparse4D v2 \cite{lin2023sparse4d} is introduced to recurrently fuse temporal information while maintaining competitive performance. Meanwhile, camera-Lidar fusion provides a strong baseline for offline detection despite its higher latency. For example, SparseLIF \cite{zhang2024sparselif} employs a lightweight 2D detector together with a monocular 3D detector to generate initial 3D queries from perspective views. These 3D queries subsequently interact with both camera and lidar features, achieving an NDS of 0.746 without relying on temporal information.

\subsection{3D Multi-object Tracking}
\begin{figure}[ht]
\centering
\includegraphics[width=\linewidth]{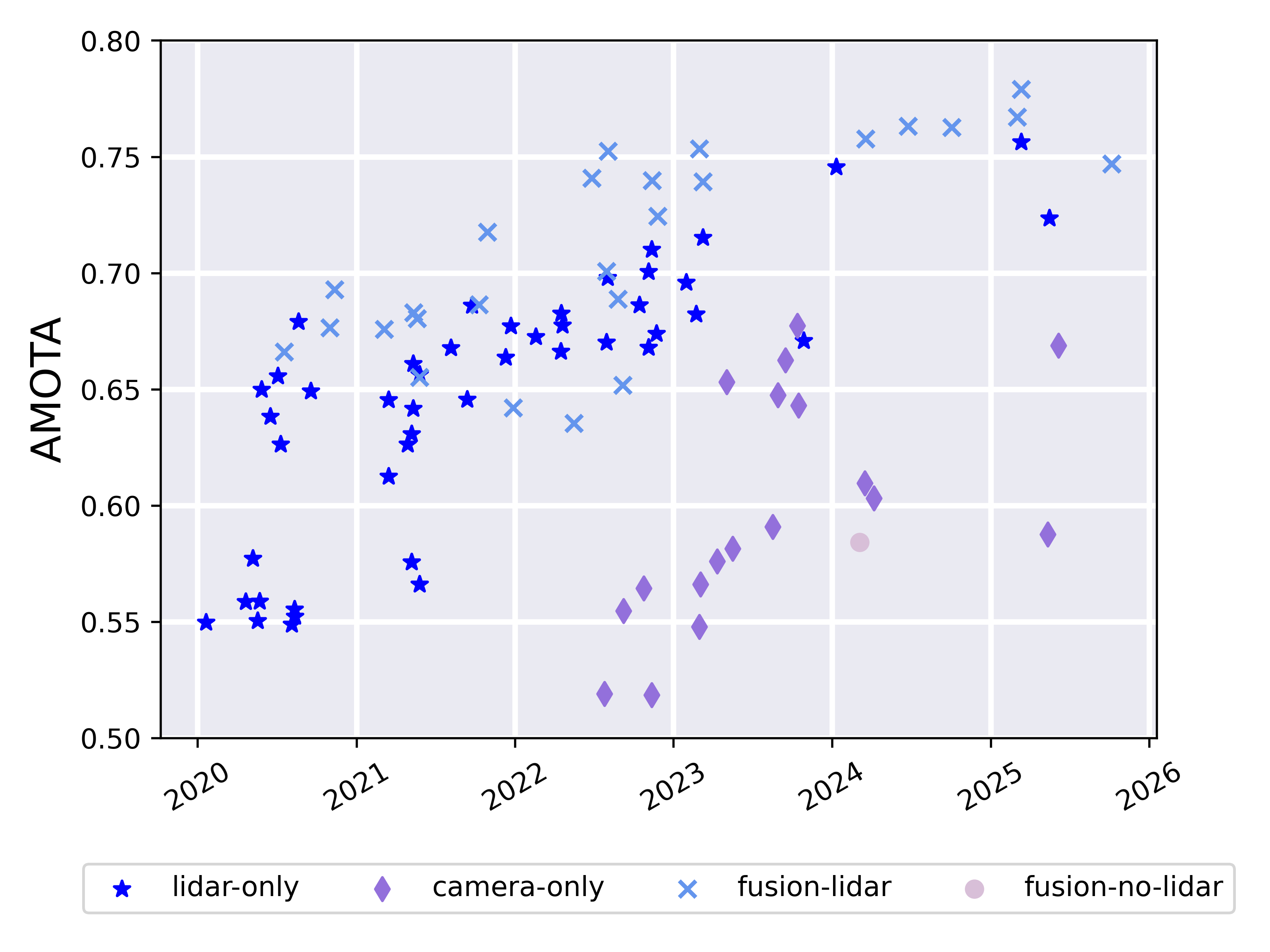}
\caption{Results on the nuScenes tracking leaderboard.}
\label{fig:challenge_tracking}
\end{figure}

Fig.~\ref{fig:challenge_tracking} shows the performance of leading methods on the 3D multi-object tracking challenge.
It shows a similar trend in AMOTA performance compared to the object detection task above, since the tracking performance largely depends on the detection performance and improvements in detection typically propagate to the tracking task.
In addition, we identify several trends in these methods.

\textbf{Transformers.} 
Transformers first emerged from the field of natural language processing, and were subsequently applied to vision tasks such as object detection.
Transformers are now being increasingly applied to 3D MOT as well.
A core process in 3DMOT is data association (DA), in which observations across historical frames are matched to produce the final tracking results.
MotionTrack~\cite{zhang2023motiontrack} formulates the DA process as the dot product process between the query, key and value matrices used in the transformer's self-attention and cross-attention mechanism.
TransMOT~\cite{ruppel2022transmot} goes beyond the DA process, and applies the transformer model to the entire tracking process, in which joint tracking and detection is done with track and object queries.

\textbf{Map information.}
Among all the tracking methods within the nuScenes community to date, only one has incorporated map information.
To improve tracking under prolonged occlusion, Offline Re-ID \cite{liu2023offlinetrackingwithobjectpermanence} uses lanes as a strong prior to estimate the motion of occluded vehicles, since the lanes guide the motion of the target vehicles. 
The use of the semantic map produces more accurate vehicle trajectories over horizons. 

\subsection{Lidar Segmentation}
\begin{figure}[ht]
\centering
\includegraphics[width=\linewidth]{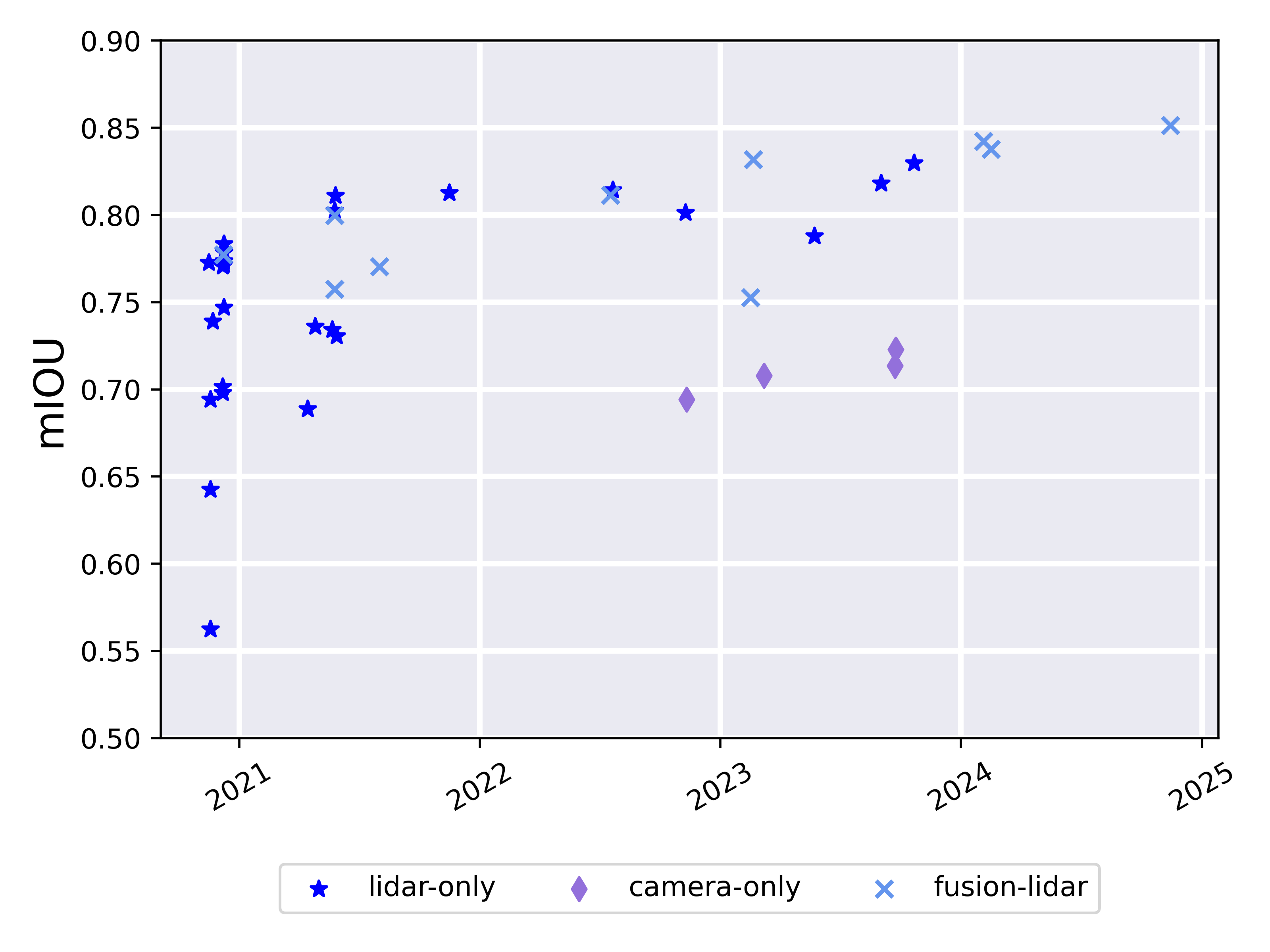}
\caption{Results on the nuScenes lidar segmentation leaderboard.
}
\label{fig:challenge_lidarseg}
\end{figure}

Fig.~\ref{fig:challenge_lidarseg} shows the performance of leading methods on the lidar segmentation challenge.
We have seen continuous interest in using the nuScenes dataset since the task was introduced. 
As seen in Fig.~\ref{fig:challenge_lidarseg}, the leading methods for lidar segmentation all use lidar or lidar fused with cameras. 
More recently there is an interest in camera-only segmentation -- wherein lidar points are classified by assigning voxel 3D semantic occupancy predictions.

In the beginning, when lidar-only was the focus of the community, most methods used point-based approaches~\cite{Pointnet,randlanet,kpconv}, which are slow and inefficient.
This transitioned to projection-based methods where point clouds are projected to a more structured and dense 2D representation which can easily be used by any 2D segmentation network.
These 2D representations can be in the form of a \textit{range-view}~\cite{rangenet++} or \textit{bird's eye view (BEV)}~\cite{polarnet}. 
This shifted to voxel-based~\cite{Cylinder3d,SPVNAS} approaches where these methods group points into voxels and then apply 3D convolutions for semantic segmentation. To make the methods efficient, these require special convolution operations to handle sparsity of the representation and/or speed-up processing. 
Currently, top performers ranges from a mixture of voxel-based, point-based, multi-view approaches at varying flavors~\cite{LidarMultiNet,udeerlvic,mseg3d,PointTransformerv3,AMVNET}.
PointTransformer v3 (PTv3) ~\cite{PointTransformerv3} is one of the top-performing methods that is point-based and manages to be computationally efficient by employing a localized patched attention mechanism to group points into non-overlapping patches, which enhances efficiency and enables the model to capture complex spatial relationships without the computational cost of global attention.
It is interesting to also note that some methods leverage temporal information across frames~\cite{SVQNet,TVSN}. 

\subsection{Lidar Panoptic Segmentation and Tracking}
Here we discuss leading methods in the lidar panoptic segmentation and tracking challenges.
Most methods are \textit{proposal-free}, following a pattern of learning rich features from a encoder and backbone and feeding them to both a segmentation head and an instance head to learn stuff and thing classes respectively~\cite{PHNet, 2D3DNet, EfficientLPS}. 
A representative method is Panoptic-PHNet~\cite{PHNet} which uses voxel and BEV encoders for feature extraction from raw sensor and KNN-transformer for learning instance centers, which are clustered to obtain instances.
More recently, multi-modal methods are also explored.  LCPS~\cite{LCPS} is the first to conduct lidar-camera fusion for panoptic tracking through novel modality-specific encodings and fusing through adding rich point-based features coming from image features. 4DFormer~\cite{waabi4dformer} also fuses lidar and camera but through a novel transformer-based architecture where it decodes instances through learnable queries and a Tracklet Association Module.

\subsection{Prediction / Motion Forecasting}
Here we discuss leading methods in the prediction challenge.
At the time of release, nuScenes stood out as a dataset for motion forecasting as it provided access to detailed HD maps.
As discussed in Section~\ref{sec:dataset_maps}, these maps contain semantic and geometric information about a wide variety of map layers and objects, such as lanes, crosswalks, stop signs, and more.
Early models~\cite{cui2019multimodal, messaoud2021trajectory, chai2019multipath} produce output trajectories by rasterizing trajectories of agents and map information as BEV images, and encoding the information using Convolutional Neural Networks.
However, rasterization of HD maps can be computationally inefficient, erase information due to occlusions. 
As an alternative, \cite{gao2020vectornet, liang2020learning, kim2021lapred, gilles2022gohome, gilles2021thomas} vectorize HD maps to use the map information more efficiently, maintain topology information and avoid rasterization artifacts.
For example, a lane boundary contains multiple control points that build a spline; a crosswalk is a polygon defined by several points; a stop sign is represented by a single point. All these geographic entities can be approximated as polylines defined by multiple control points, along with their attributes. 
Neighboring entities can even share control points.
Similarly, the dynamics of moving agents can also be approximated by polylines based on their motion trajectories. All these polylines can then be represented as sets of vectors, and encoded by Graph Neural Networks for trajectory/polyline features. 
The vector representations are able to attain good performance while using fewer parameters, compared to rasterization-based approaches.

\subsection{Other benchmarks}
Beyond the official nuScenes benchmarks presented in the previous sections, we present further community-organized benchmarks using the nuScenes dataset.

\textbf{2D detection and segmentation in nuImages.} 
nuImages complements nuScenes in providing 2D annotated image data, with instance and segmentation masks of various classes, to expedite AV research on 2D object detection, monocular 3D object detection, and image-based semantic or instance segmentation. 
Due to a lack of official train/val splits, the community uses the splits provided by MMDetection3D~\cite{mmdet3d2020}.

Similar to nuScenes, nuImages is collected in diverse locations, lightning and weather conditions making it very challenging. In fact, several research have leveraged this for Out-of-Domain object detection research~\cite{ODD,OutlierAwareObjectDetection}.

Because nuScenes and nuImages were both collected in similar sensor set-ups and conditions, they have the same Operating Domain.
This means that nuImages can help with image-based research in nuScenes, specifically camera-based methods can use nuImages as a dataset for pretraining.
For example,~\cite{PointPainting, Frustumformer} use the dataset as pretraining for 3D detection,~\cite{Sparse4d, StreamPETR, SSLtemporalordering} also uses it as pretraining, but also leverage the temporal information from nuImages as mentioned in Section~\ref{sec:dataset_extensions}.
It is also used as extra priors for learning auxiliary tasks~\cite{2D3DNet, pseudolabels}.
In conjunction, nuImages also provides the ego-vehicle pose and calibrated sensor information which unlocks new tasks such as ego-vehicle speed estimation~\cite{EgoSpeed} at scale.
Finally, we note that nuImages provides attributes for foreground objects.
While attributes like flashing emergency vehicle lights and sitting pedestrians are helpful for AVs, we are not aware of any studies that learn to predict these attributes.
Other research that leverages nuImages includes active learning~\cite{ActiveLearning} and few-shot object detection~\cite{fewshotobjdetection}.

\textbf{Map Element Detection.} 
As presented in Section~\ref{sec:dataset_maps}, HD maps are crucial for autonomous driving so that the vehicle can localize itself accurately with respect to the 3D world, and other modules such as \textit{planning}, \textit{prediction} and \textit{perception} can utilize map semantic priors to be robust in route planning, object detection, and motion forecasting.
However, building HD maps requires laborious human annotations which can be costly and not scalable. On top of that, environmental changes require continuous updates and maintenance of maps. 
In this light, several efforts have focused on working on online mapping, which builds the map on the fly. 

With the full sensor suite and the availability of semantic maps, nuScenes has become a default benchmark dataset for the community.
Research efforts differ from performing BEV semantic segmentation~\cite{liftsplatshoot, PyOccNet} to directly creating vectorized map objects~\cite{instagram,hdmapnet,vectormapnet,MapTR,Pivotnet,Maptrv2}.
The former is a simpler task, but traditional downstream modules require many heuristics to build the vectorized polyline instances from rasterized images.
The latter is a more challenging task, but can be trained end-to-end and is more directly needed downstream. 
This can be further expanded for \textit{full} end-to-end driving where we learn future waypoints of the ego-vehicle directly. 
The BEV semantic segmentation task is based on multiple cameras, learning both static road elements and dynamic objects. Before, annotations were scarce and the community relied on synthetic data, stereo or sim2real data to transform camera images to BEV space accurately~\cite{VPN, VED, CAM2BEV, Learning2look}. But with nuScenes, direct supervision from map layers is possible.  Pioneering work on BEV semantic segmentation in nuScenes is LiftSplatShoot~\cite{liftsplatshoot} and PyOccNet~\cite{PyOccNet} which succeeding works followed~\cite{Hft, STA, Trans2maps, Gitnet}. Techniques vary from fully-connected bottleneck layers~\cite{VPN, VED, PYVA, PyOccNet, OFT}, leveraging depth information~\cite{EPOSH, STA, liftsplatshoot} or temporal information~\cite{STA} or attention mechanism~\cite{Gitnet}.

HDMapNet~\cite{hdmapnet} is a formative work in the task of building online vectorized HD maps using nuScenes. However, it still requires a two-step process of predicting intermediate forms and post-processing these to estimate polylines and polygons. 
VectorMapNet~\cite{vectormapnet} does this end-to-end.
Subsequent works~\cite{MapTR, Maptrv2, Pivotnet, bimapnet} use the same pattern of first learning BEV representations using surround cameras and/or lidar, followed by a transformer-based map-element decoder, with some works of adding temporal queries~\cite{Yuan_2024_streammapnet}
This is similar to the latest works on 3D object detection but has a more complex decoding strategy due to the increased output dimensions and variety of shapes for polylines and polygons. 

\textbf{Semantic Occupancy Prediction.}
We organize challenges\footnote{The occupancy prediction challenges are co-organized with OpenDriveLab.} for semantic occupancy prediction, in which the task is to predict the 3D occupancy of a given scene, and the flow of the objects belonging to the thing classes.
This is an image-based task, in which only the images from the six cameras are used (no lidar or radar modalities are involved).
Participants are scored based on the Occupancy Score (OccScore) metric, which consists of two components: 
1) ray-based mIOU~\cite{liu2023fullysparse3dpanopticoccupancyprediction} for occupancy, and 
2) absolute velocity error (AVE) for flow. 
In ray-based mIOU, rays are projected into the predicted 3D occupancy volume and the class label and flow prediction at the surface at which each ray intersects is compared against the ground truth.
For AVE, a threshold of 2 meters is used and is computed for 8 classes (car, truck, trailer, bus, construction vehicle, bicycle, motorcycle and pedestrian).
The Occupancy Score is then the weighted sum of mIOU and mAVE.

\textbf{Planning Challenge}
Although nuScenes does not officially support the planning task, it has become a default dataset for open-loop planning~\cite{rethinkingopenloop} using the evaluation protocol of L2 errors and collision rate~\cite{hu2022st, hu2023planning, jiang2023vad}.

In~\cite{rethinkingopenloop} the authors discovered that a planning method that uses only the past and present ego state can achieve state-of-the-art performance, outperforming even methods that rely on camera, lidar or map data. This exposes the fundamental flaws of the open-loop evaluation, which is not specific to this dataset. 
The effects may however have been exacerbated by the short planning horizon (3s) and relatively small dataset size in the unofficial planning evaluation protocol of nuScenes.
\cite{li2023egostatus} further investigate why end-to-end planning methods on nuScenes rely primarily on the vehicle's ego state, rather than perception inputs.
To address this issue, they propose improved open loop metrics for collisions with other agents and intersections with the road boundary.
Ultimately we believe that the limitations of open loop training and evaluation are well known and that the community should move towards closed loop simulation.
The follow-up dataset nuPlan~\cite{caesar2022nuplan,karnchanachari24nuplan} addresses this gap.
It is approximately 235x larger and includes a framework for closed-loop evaluation of planners.

\section{Discussion}
In this section we discuss some of the strengths and weaknesses of the nuScenes dataset.
These are based on the literature, user feedback and our own observations.

\textbf{Domain shift.}
We analyze the performance of several object detection methods in the literature.
Specifically, we compare the val and test split performance gaps.
While data should not leak from the trainval to the test split, it is also important that the test set follows a similar distribution to the trainval split, as otherwise leaderboard rankings will be obfuscated by random fluctuations.
We conduct an analysis of the performance gaps of 32 different 3D object detection methods on 3 datasets: nuScenes, Waymo~\cite{sun2019wod} and KITTI~\cite{geiger2012kitti}. 
We find that the median mAP performance gap between results on the val and test split is 1.6\% for nuScenes, 3.8\% for Waymo and 3.6\% for KITTI.
While the metrics and evaluation protocols are not directly comparable, this seems to imply that domain shifts are less of a problem in nuScenes.
When ranking different methods in a benchmark, it is particularly important that the ranking is heavily correlated between val and test.
We compute the Pearson correlation coefficient of the above methods and find a correlation of 99.7\% for nuScenes, 89.8\% for Waymo and 82.4\% for KITTI, again indicating that the val and test splits of nuScenes are more similar.

\begin{figure} 
    \centering
    \includegraphics[width=\linewidth]{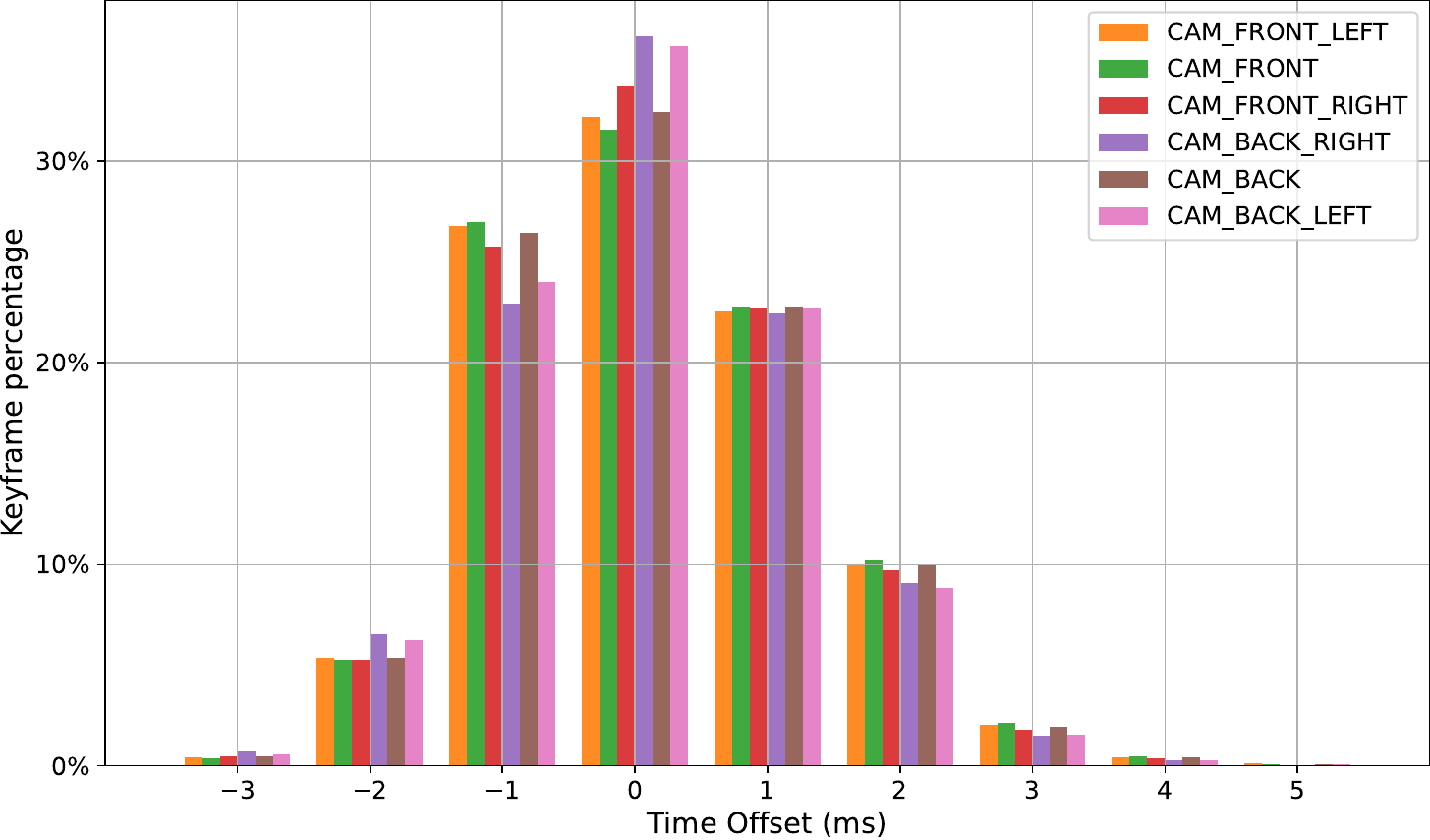}
  \caption{A histogram of the camera-lidar time offsets in nuScenes. 
  The synchronization is of high quality, since we smartly select keyframes and discard scenes that do not meet our quality requirements.
  }
  \label{fig:camera_lidar_sync_stats}
\end{figure}

\textbf{Sensor synchronization.}
As discussed in Sec.~\ref{sec:dataset-postprocessing}, the nuScenes dataset follows several steps to make sure that the synchronization of lidar and camera sensors is within the specified quality requirements.
Some users have reported issues with the synchronization in nuScenes~\cite{nuscenes-misalignment}.
Upon further inspection of provided examples we usually identify misunderstandings about parallax to be the primary cause of the misalignment issues.
Namely, due to the different viewpoints of camera and lidar, the hypothetical lines through points from multiple lidar beams can intersect, which they would not if viewed from the perspective of the lidar sensor.
In some cases we observe an intricate interplay between multiple factors such as synchronization, calibration, lack of 6DOF localization and parallax that can make it difficult to identify the root cause.
To isolate the synchronization aspect, we compare the camera and lidar timestamps in nuScenes.
Since the lidar is timestamped at the beginning of a rotation and not when it is facing the same yaw angle as the camera, we follow the practice of~\cite{sun2019wod} to adjust the timestamp proportionally to the yaw orientation of the camera, thus assuming a constant angular velocity of the lidar.
We then compute the \emph{time offset} as the difference between the timestamps of the camera and the lidar.
However, since we do not know the exact yaw orientation at the beginning of a lidar scan, we subtract from the time offsets the median time offset for each camera across the entire dataset.
Fig.~\ref{fig:camera_lidar_sync_stats} shows a histogram of the camera-lidar time offsets as computed by this method.
We observe with at least 99.7\% confidence the time offsets are bounded in $[-4.5, 5.0]$ ms.
In contrast, for WOD~\cite{sun2019wod} the bounds are $[-6, 7]$ ms.
We can thus infer that the synchronization of nuScenes is slightly better than in WOD and significantly better than the quality requirement of $15$ ms that we defined in Sec.~\ref{sec:dataset-postprocessing}.
To put this in perspective, a time offset of $1$ ms between camera and lidar corresponds to the ego vehicle moving $1$ cm at a speed of $10$ m/s. In a static scene, this means a lidar point could appear shifted by about $1$ cm relative to the camera frame. This displacement is typically within the positional accuracy margin of automotive lidar sensors, which is usually on the order of a few centimeters.

\textbf{Scenes vs. logs.}
As described in Section~\ref{sec:dataset}, the nuScenes dataset consists of 1000 scenes (5.5h), carefully curated from 83 longer driving sequences (14.8h).
Splitting the dataset into a large number of shorter scenes has many advantages over datasets with a smaller number of long driving sequences~\cite{geiger2012kitti}.
First, the scenes are manually curated and thus we focus on interesting scenarios, such as dense traffic at intersections. 
A downside is that through manual curation, active learning research~\cite{wei2024basal} is inhibited in its effectiveness, since less informative scenes are already discarded.
Second, for a large number of scenes it becomes much easier to create balanced train, val and test splits that follow the same data distribution.
We hypothesize that this is one reason why there is a lower test-val rank correlation on KITTI (22 sequences) than on nuScenes (1000 scenes), as described above.
In fact, the trainval and test logs in nuScenes are created by evaluating numerous candidate split assignments and selecting the one that best balances several criteria (night/rain/regular conditions, Singapore vs. Boston).

\textbf{Geographically overlapping data splits.}
In nuScenes the train, val and test splits are collected from the same regions. 
This design choice was made due to the focus on a robotaxi use-case in geofenced areas, rather than ADAS applications that need to generalize to any environment.
For perception and planning tasks, where foreground objects are more important, it is acceptable and desirable to overfit for a particular environment.
For mapping tasks, where background objects are more important, this type of overfitting is undesirable.
An example for this is~\cite{li2023toponet}, which models lane topology.
They observe that their network attention is overfitting to patterns on building, rather than focusing on the road that defines the lane topology. 
To address this issue~\cite{PyOccNet} created new non-overlapping splits for monocular BEV semantic mapping, while preserving a similar data distribution.
Similarly,~\cite{lilja2024dataleakage} also create new splits for mapping purposes, noting that ``over 80\% of nuScenes and 40\% of Argoverse 2 validation and test samples are less than 5m from a training sample''. 
Many works in the literature still resort to the original nuScenes splits not designed for mapping tasks, which we discourage.

\textbf{Urban vs. highway.}
The nuScenes dataset includes urban and suburban driving environments in Singapore and Boston, which are representative of early robotaxi operations at sub-highway speeds.
While ADAS development has traditionally focused on these environments, the recent deployment of commercial robotaxi services on highways is shifting attention toward higher-speed scenarios. 
Highway driving~\cite{comma-2k19} is often considered easier in terms of traffic complexity, but higher speeds increase the risks and impose more stringent requirements on perception systems.
Two major challenges in highway environments are adverse weather and long-range perception
Adverse weather conditions are typically addressed by including radar sensors~\cite{aimotive,zenseact,KRadar,radial} which operate at longer wavelengths than lidar and can therefore penetrate rain, fog, or snow more effectively.
Recently more datasets focus on long range perception. 
Notably, radar in lidar in~\cite{zenseact} have a range of up to 250m, while~\cite{aevascenes} uses the first Frequency-Modulated Continuous Wave (FMCW) lidar and achieves a range of 400m.

\textbf{Missing 6DOF localization.} 
While our geometric map does have 3D information, we do not expose the z information in our annotations nor provide a ground height layer; hence projecting them to the camera frame can result in misalignments. To this date, the only dataset to provide this information is Argoverse 2, albeit only at a 30cm resolution.
In addition, localization is done in the 2D raster-level, which makes the estimated pose inaccurate.

\begin{figure}[ht]
\centering
\includegraphics[width=\linewidth]{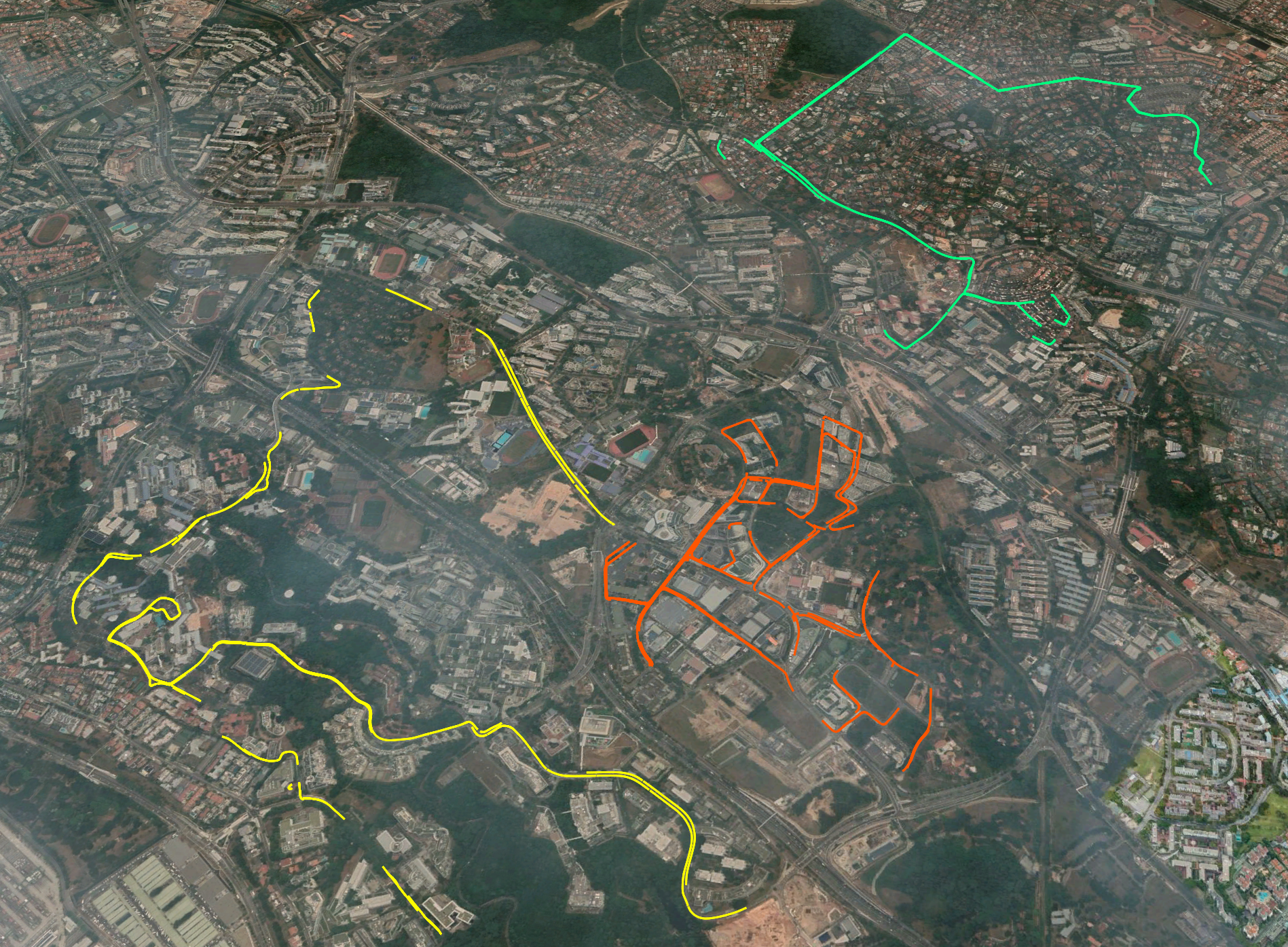}
\caption{Alignment of nuScenes maps in Singapore with a satellite image that is registered to the global map coordinate system.}
\label{fig:global_alignment}
\end{figure}

\begin{figure}[ht]
\centering
\includegraphics[width=\linewidth, trim=0 0 0 7cm, clip]{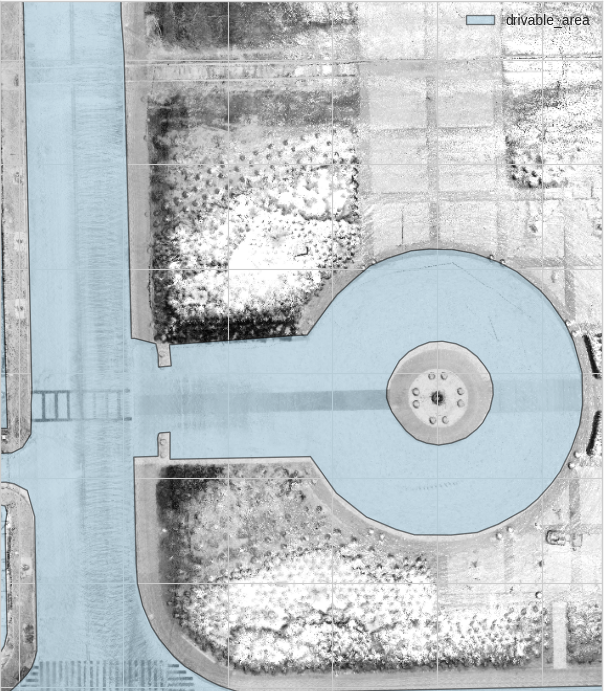}
\caption{Minor misalignment between drivable area mask and the intensity basemap in nuScenes.}
\label{fig:bad_alignment}
\end{figure}

\textbf{Map misalignment.}
We do several runs of mapping data collection across different routes to obtain a city-scale maps. 
Geometric maps from different collections are merged via optimizing the matching factors and loop closure factors. 
This leads to small non-linear transformations on the merged areas. Semantic map polygon annotations are also combined by relying on the transformation shifts obtained from the geometric maps. 
These subsequent transformations lead to a lack of global registrations in our maps. 
This can cause misalignment of our maps when projecting with the global map coordinate system. 
Hence, careful visualization is required when using external map information with our data and projection to our basemaps. Fig.~\ref{fig:global_alignment} shows an example where our map elements are visually verified to be correct when projecting to a global coordinate system. Fig.~\ref{fig:bad_alignment} shows known misalignments due to inaccurate transformation shifts in the semantic layers. 

\section{Conclusion}
\label{sec:conclusion}
The nuScenes dataset has had a significant impact on the field of Autonomous Vehicles.
nuScenes builds upon the multimodal nature of the KITTI~\cite{geiger2012kitti} dataset, but extends the hardware setup to a complete sensor suite that was used for the world's first self-driving taxi service.
In this paper we summarize in detail how the nuScenes dataset was created, sharing hitherto unpublished details.
We critically review many of its aspects and give suggestions how to further improve upon these.
Finally, we discuss the various official and unofficial benchmark tasks on the nuScenes dataset and sketch the development of the leading methods on these benchmarks over time.
The next generation dataset by Motional, nuPlan~\cite{karnchanachari24nuplan}, goes beyond perception and prediction and features a closed-loop simulation framework for machine learning-based planning and a vastly larger dataset.
Future dataset generations should focus even more on rare corner-case scenarios, both in terms of perception and planning, which currently are responsible for the majority of autonomous driving accidents.
\ifCLASSOPTIONcompsoc
  \section*{Acknowledgments}
\else
  \section*{Acknowledgment}
\fi

The authors would like to thank
Shiming Wang for analyzing the synchronization quality and domain shifts on different datasets,
Giancarlo Baldan for support and guidance,
Andras Palffy for his expertise with radar,
Kashyap Chitta for highlighting the challenges with open loop planning evaluation.
Ishay Beery for summarizing the developments in tracking,
Alexander Wischnewski for highlighting the lack of highway datasets,
Jannik Zuern for discussions on lane graph estimation,
as well as Alexander Liniger, Sima Chonghao, Li Chen and Junsheng Fu for discussions on cross-area generalization in mapping tasks.




 
%
{
\bibliographystyle{IEEEtran}
\bibliography{paper}

@ARTICLE{pseudolabels,
  author={Liu, Cheng},
  journal={IEEE Access}, 
  title={Enhance the 3D Object Detection With 2D Prior}, 
  year={2024}
}

@misc{aevascenes,
    title        = {AevaScenes: A Dataset and Benchmark for FMCW LiDAR Perception},
    author       = {Narasimhan, Gautham Narayan and Vhavle, Heethesh and Vishvanatha, Kumar Bhargav and Reuther, James},
    year         = {2025}
  }

@InProceedings{
Kan_2024_icra, 
author={Chen, Kan and Ge, Runzhou and Qiu, Hang and Ai-Rfou, Rami and Qi, Charles R. and Zhou, Xuanyu and Yang, Zoey and Ettinger, Scott and Sun, Pei and Leng, Zhaoqi and Mustafa, Mustafa and Bogun, Ivan and Wang, Weiyue and Tan, Mingxing and Anguelov, Dragomir}, 
title={WOMD-LiDAR: Raw Sensor Dataset Benchmark for Motion Forecasting}, 
month={May}, 
booktitle={ICRA},
year={2024} }

@inproceedings{StreamPETR,
  title={Exploring Object-Centric Temporal Modeling for Efficient Multi-View 3D Object Detection},
  author={Wang, Shihao and Liu, Yingfei and Wang, Tiancai and Li, Ying and Zhang, Xiangyu},
  booktitle={ICCV},
  year={2023}
}

@inproceedings{frustumformer,
  title={Frustumformer: Adaptive instance-aware resampling for multi-view 3d detection},
  author={Wang, Yuqi and Chen, Yuntao and Zhang, Zhaoxiang},
  booktitle={CVPR},
  year={2023}
}

@article{SSLtemporalordering,
  title={Self-supervised representation learning from temporal ordering of automated driving sequences},
  author={Lang, Christopher and Braun, Alexander and Schillingmann, Lars and Haug, Karsten and Valada, Abhinav},
  journal={RA-L},
  year={2024}
}

@article{Sparse4d,
  title={Sparse4d: Multi-view 3d object detection with sparse spatial-temporal fusion},
  author={Lin, Xuewu and Lin, Tianwei and Pei, Zixiang and Huang, Lichao and Su, Zhizhong},
  journal={arXiv preprint arXiv:2211.10581},
  year={2022}
}

@article{ODD,
  title={Efficient out-of-distribution detection using latent space of $\beta$-vae for cyber-physical systems},
  author={Ramakrishna, Shreyas and Rahiminasab, Zahra and Karsai, Gabor and Easwaran, Arvind and Dubey, Abhishek},
  journal={TCPS},
  volume={6},
  number={2},
  pages={1--34},
  year={2022},
  publisher={ACM New York, NY}
}

@inproceedings{OutlierAwareObjectDetection,
  title={Normalizing flow based feature synthesis for outlier-aware object detection},
  author={Kumar, Nishant and {\v{S}}egvi{\'c}, Sini{\v{s}}a and Eslami, Abouzar and Gumhold, Stefan},
  booktitle={CVPR},
  year={2023}
}

@inproceedings{ActiveLearning,
  title={Multi-task consistency for active learning},
  author={Hekimoglu, Aral and Friedrich, Philipp and Zimmer, Walter and Schmidt, Michael and Marcos-Ramiro, Alvaro and Knoll, Alois},
  booktitle={ICCV},
  year={2023}
}

@article{fewshotobjdetection,
  title={Revisiting few-shot object detection with vision-language models},
  author={Madan, Anish and Peri, Neehar and Kong, Shu and Ramanan, Deva},
  journal={NeurIPS Datasets and Benchmark Track},
  year={2024}
}

@article{Menet,
  title={MENet: Map-enhanced 3D object detection in bird’s-eye view for LiDAR point clouds},
  author={Huang, Yuanxian and Zhou, Jian and Li, Xicheng and Dong, Zhen and Xiao, Jinsheng and Wang, Shurui and Zhang, Hongjuan},
  journal={International Journal of Applied Earth Observation and Geoinformation},
  year={2023}
}

@inproceedings{RadarRobotCarDatasetICRA2020,
  address = {Paris},
  author = {Dan Barnes and Matthew Gadd and Paul Murcutt and Paul Newman and Ingmar Posner},
  title = {The Oxford Radar RobotCar Dataset: A Radar Extension to the Oxford RobotCar Dataset},
  booktitle = {ICRA},
  year = {2020}
}

@article{delft,
  title={Delft: Views on Delft},
  author={Tanis, F and Sioli, A and Stani{\v{c}}i{\'c}, Aleksandar and Havik, KM and Dale, HE and Vogel, WC and de Wit, SI and Hernandez, JA Mejia and Guembe, E P{\'e}rez and Pllumbi, Dorina and others},
  journal={Writingplace: Journal for Architecture and Literature},
  number={8-9},
  pages={311--331},
  year={2023},
}

@InProceedings{Yuan_2024_streammapnet,
    author    = {Yuan, Tianyuan and Liu, Yicheng and Wang, Yue and Wang, Yilun and Zhao, Hang},
    title     = {StreamMapNet: Streaming Mapping Network for Vectorized Online HD Map Construction},
    booktitle = {WACV},
    year      = {2024},
}

@inproceedings{bimapnet,
  title={End-to-end vectorized hd-map construction with piecewise bezier curve},
  author={Qiao, Limeng and Ding, Wenjie and Qiu, Xi and Zhang, Chi},
  booktitle={CVPR},
  year={2023}
}

@INPROCEEDINGS{LaneFusion,
  author={Fujimoto, Taisei and Tanaka, Satoshi and Kato, Shinpei},
  booktitle={IV}, 
  title={LaneFusion: 3D Object Detection with Rasterized Lane Map}, 
  year={2022}
}

@inproceedings{SVQNet,
  title={{SVQNet}: Sparse voxel-adjacent query network for 4d spatio-temporal lidar semantic segmentation},
  author={Chen, Xuechao and Xu, Shuangjie and Zou, Xiaoyi and Cao, Tongyi and Yeung, Dit-Yan and Fang, Lu},
  booktitle={ICCV},
  year={2023}
}

@InProceedings{LCPS,
    author    = {Zhang, Zhiwei and Zhang, Zhizhong and Yu, Qian and Yi, Ran and Xie, Yuan and Ma, Lizhuang},
    title     = {LiDAR-Camera Panoptic Segmentation via Geometry-Consistent and Semantic-Aware Alignment},
    booktitle = {ICCV},
    year      = {2023}
}

@article{fent2024man,
  title={MAN TruckScenes: A multimodal dataset for autonomous trucking in diverse conditions},
  author={Fent, Felix and Kuttenreich, Fabian and Ruch, Florian and Rizwin, Farija and Juergens, Stefan and Lechermann, Lorenz and Nissler, Christian and Perl, Andrea and Voll, Ulrich and Yan, Min and others},
  journal={arXiv preprint arXiv:2407.07462},
  year={2024}
}

@InProceedings{waabi4dformer,
    title = {4D-Former: Multimodal 4D Panoptic Segmentation},
    author = {Athar, Ali and Li, Enxu and Casas, Sergio and Urtasun, Raquel},
    booktitle = {CoRL},
    year = {2023},
  }

@inproceedings{kpconv,
  title={Kpconv: Flexible and deformable convolution for point clouds},
  author={Thomas, Hugues and Qi, Charles R and Deschaud, Jean-Emmanuel and Marcotegui, Beatriz and Goulette, Fran{\c{c}}ois and Guibas, Leonidas J},
  booktitle={CVPR},
  year={2019}
}

@inproceedings{Pointnet,
  title={Pointnet: Deep learning on point sets for 3d classification and segmentation},
  author={Qi, Charles R and Su, Hao and Mo, Kaichun and Guibas, Leonidas J},
  booktitle={CVPR},
  year={2017}
}

@inproceedings{randlanet,
  title={Randla-net: Efficient semantic segmentation of large-scale point clouds},
  author={Hu, Qingyong and Yang, Bo and Xie, Linhai and Rosa, Stefano and Guo, Yulan and Wang, Zhihua and Trigoni, Niki and Markham, Andrew},
  booktitle={CVPR},
  year={2020}
}

@article{TVSN,
  title={Learning spatial and temporal variations for 4D point cloud segmentation},
  author={Hanyu, Shi and Jiacheng, Wei and Hao, Wang and Fayao, Liu and Guosheng, Lin},
  journal={arXiv preprint arXiv:2207.04673},
  year={2022}
}

@inproceedings{PointTransformerv3,
  title={Point Transformer V3: Simpler Faster Stronger},
  author={Wu, Xiaoyang and Jiang, Li and Wang, Peng-Shuai and Liu, Zhijian and Liu, Xihui and Qiao, Yu and Ouyang, Wanli and He, Tong and Zhao, Hengshuang},
  booktitle={CVPR},
  year={2024}
}

@inproceedings{mseg3d,
  title={Mseg3d: Multi-modal 3d semantic segmentation for autonomous driving},
  author={Li, Jiale and Dai, Hang and Han, Hao and Ding, Yong},
  booktitle={CVPR},
  year={2023}
}

@article{udeerlvic,
  title={{LVIC}: Multi-modality segmentation by Lifting Visual Info as Cue},
  author={Dong, Zichao and Pang, Bowen and Huang, Xufeng and Ji, Hang and Zhan, Xin and Chen, Junbo},
  journal={arXiv preprint arXiv:2403.05159},
  year={2024}
}

@inproceedings{culane,
 author = {Xingang Pan and Jianping Shi and Ping Luo and Xiaogang Wang and and Xiaoou Tang},
 title = {Spatial As Deep: Spatial CNN for Traffic Scene Understanding},
 booktitle = {AAAI},
 year = {2018} 
}

@INPROCEEDINGS{Lanelet2,
  title     = {Lanelet2: A High-Definition Map Framework for the Future of Automated Driving},
  author    = {Poggenhans, Fabian and Pauls, Jan-Hendrik and Janosovits, Johannes and Orf, Stefan and Naumann, Maximilian and Kuhnt, Florian and Mayr, Matthias},
  booktitle = {ITSC},
  year      = {2018}
}

@INPROCEEDINGS{Lanelet2fornuScenes,
  title={Lanelet2 for nuScenes: Enabling Spatial Semantic Relationships and Diverse Map-Based Anchor Paths},
  author={Naumann, Alexander and Hertlein, Felix and Grimm, Daniel and Zipfl, Maximilian and Thoma, Steffen and Rettinger, Achim and Halilaj, Lavdim and Luettin, Juergen and Schmid, Stefan and Caesar, Holger},
  booktitle={CVPR},
  year={2023}
}

@inproceedings{Honda3D,
    author = {Abhishek Patil and Srikanth Malla and Haiming Gang and Yi-Ting Chen},
    title = {The {H3D} Dataset for Full-Surround 3D Multi-Object Detection and Tracking in Crowded Urban Scenes},  
    booktitle = {ICRA},
    year = {2019}
}

@inproceedings{chai2019multipath,
  title={Multipath: Multiple probabilistic anchor trajectory hypotheses for behavior prediction},
  author={Chai, Yuning and Sapp, Benjamin and Bansal, Mayank and Anguelov, Dragomir},
  booktitle={CoRL},
  year={2019}
}

@inproceedings{messaoud2021trajectory,
  title={Trajectory prediction for autonomous driving based on multi-head attention with joint agent-map representation},
  author={Messaoud, Kaouther and Deo, Nachiket and Trivedi, Mohan M and Nashashibi, Fawzi},
  booktitle={IV},
  year={2021}
}

@inproceedings{cui2019multimodal,
  title={Multimodal trajectory predictions for autonomous driving using deep convolutional networks},
  author={Cui, Henggang and Radosavljevic, Vladan and Chou, Fang-Chieh and Lin, Tsung-Han and Nguyen, Thi and Huang, Tzu-Kuo and Schneider, Jeff and Djuric, Nemanja},
  booktitle={ICRA},
  year={2019}
}

@article{choi2018kaist,
  title={KAIST multi-spectral day/night data set for autonomous and assisted driving},
  author={Choi, Yukyung and Kim, Namil and Hwang, Soonmin and Park, Kibaek and Yoon, Jae Shin and An, Kyounghwan and Kweon, In So},
  journal={ITSC},
  year={2018}
}

@inproceedings{ma2019trafficpredict,
  title={Trafficpredict: Trajectory prediction for heterogeneous traffic-agents},
  author={Ma, Yuexin and Zhu, Xinge and Zhang, Sibo and Yang, Ruigang and Wang, Wenping and Manocha, Dinesh},
  booktitle={AAAI},
  year={2019}
}

@INPROCEEDINGS{Openlanev2,
  title={OpenLane-V2: A Topology Reasoning Benchmark for Unified 3D {HD} Mapping}, 
  author={Wang, Huijie and Li, Tianyu and Li, Yang and Chen, Li and Sima, Chonghao and Liu, Zhenbo and Wang, Bangjun and Jia, Peijin and Wang, Yuting and Jiang, Shengyin and Wen, Feng and Xu, Hang and Luo, Ping and Yan, Junchi and Zhang, Wei and Li, Hongyang},
  booktitle={NeurIPS},
  year={2023}
}

@INPROCEEDINGS {argoverse1,
  author = {Ming-Fang Chang and John W Lambert and Patsorn Sangkloy and Jagjeet Singh
       and Slawomir Bak and Andrew Hartnett and De Wang and Peter Carr
       and Simon Lucey and Deva Ramanan and James Hays},
  title = {Argoverse: 3D Tracking and Forecasting with Rich Maps},
  booktitle = {CVPR},
  year = {2019}
}

@inproceedings{gao2020vectornet,
  title={Vectornet: Encoding {HD} maps and agent dynamics from vectorized representation},
  author={Gao, Jiyang and Sun, Chen and Zhao, Hang and Shen, Yi and Anguelov, Dragomir and Li, Congcong and Schmid, Cordelia},
  booktitle={CVPR},
  year={2020}
}

@inproceedings{liang2020learning,
  title={Learning lane graph representations for motion forecasting},
  author={Liang, Ming and Yang, Bin and Hu, Rui and Chen, Yun and Liao, Renjie and Feng, Song and Urtasun, Raquel},
  booktitle={ECCV},
  year={2020}
}

@inproceedings{gilles2022gohome,
  title={Gohome: Graph-oriented heatmap output for future motion estimation},
  author={Gilles, Thomas and Sabatini, Stefano and Tsishkou, Dzmitry and Stanciulescu, Bogdan and Moutarde, Fabien},
  booktitle={ICRA},
  year={2022}
}

@inproceedings{gilles2021thomas,
  title={{THOMAS}: Trajectory heatmap output with learned multi-agent sampling},
  author={Gilles, Thomas and Sabatini, Stefano and Tsishkou, Dzmitry and Stanciulescu, Bogdan and Moutarde, Fabien},
  booktitle={ICLR},
  year={2022}
}

@inproceedings{kim2021lapred,
  title={Lapred: Lane-aware prediction of multi-modal future trajectories of dynamic agents},
  author={Kim, ByeoungDo and Park, Seong Hyeon and Lee, Seokhwan and Khoshimjonov, Elbek and Kum, Dongsuk and Kim, Junsoo and Kim, Jeong Soo and Choi, Jun Won},
  booktitle={CVPR},
  year={2021}
}

@ARTICLE{KRadar,
  title={K-radar: 4d radar object detection for autonomous driving in various weather conditions},
  author={Paek, Dong-Hee and Kong, Seung-Hyun and Wijaya, Kevin Tirta},
  journal={NeurIPS},
  year={2022}
}

@INPROCEEDINGS{zenseact,
  title={Zenseact Open Dataset: A large-scale and diverse multimodal dataset for autonomous driving},
  author={Alibeigi, Mina and Ljungbergh, William and Tonderski, Adam and Hess, Georg and Lilja, Adam and Lindstr{\"o}m, Carl and Motorniuk, Daria and Fu, Junsheng and Widahl, Jenny and Petersson, Christoffer},
  booktitle={ICCV},
  year={2023}
}

@INPROCEEDINGS{Pivotnet,
  title={Pivotnet: Vectorized pivot learning for end-to-end {HD} map construction},
  author={Ding, Wenjie and Qiao, Limeng and Qiu, Xi and Zhang, Chi},
  booktitle={ICCV},
  year={2023}
}

@INPROCEEDINGS{MapTR,
  title={MapTR: Structured Modeling and Learning for Online Vectorized {HD} Map Construction},
  author={Liao, Bencheng and Chen, Shaoyu and Wang, Xinggang and Cheng, Tianheng and Zhang, Qian and Liu, Wenyu and Huang, Chang},
  booktitle={ICLR},
  year={2023}
}

@ARTICLE{Maptrv2,
  title={MapTRv2: An End-to-End Framework for Online Vectorized {HD} Map Construction},
  author={Liao, Bencheng and Chen, Shaoyu and Zhang, Yunchi and Jiang, Bo and Zhang, Qian and Liu, Wenyu and Huang, Chang and Wang, Xinggang},
  journal={IJCV},
  year={2024}
}

@INPROCEEDINGS{OpenLane,
  title={Persformer: 3d lane detection via perspective transformer and the openlane benchmark},
  author={Chen, Li and Sima, Chonghao and Li, Yang and Zheng, Zehan and Xu, Jiajie and Geng, Xiangwei and Li, Hongyang and He, Conghui and Shi, Jianping and Qiao, Yu and others},
  booktitle={ECCV},
  year={2022}
}

@INPROCEEDINGS{Bdd100k,
  title={{BDD100K}: A diverse driving dataset for heterogeneous multitask learning},
  author={Yu, Fisher and Chen, Haofeng and Wang, Xin and Xian, Wenqi and Chen, Yingying and Liu, Fangchen and Madhavan, Vashisht and Darrell, Trevor},
  booktitle={CVPR},
  year={2020}
}

@ARTICLE{Apolloscape,
  title={The apolloscape open dataset for autonomous driving and its application},
  author={Huang, Xinyu and Wang, Peng and Cheng, Xinjing and Zhou, Dingfu and Geng, Qichuan and Yang, Ruigang},
  journal={TPAMI},
  year={2019}
}

@INPROCEEDINGS{GenLanenet,
  title={Gen-lanenet: A generalized and scalable approach for 3d lane detection},
  author={Guo, Yuliang and Chen, Guang and Zhao, Peitao and Zhang, Weide and Miao, Jinghao and Wang, Jingao and Choe, Tae Eun},
  booktitle={ECCV},
  year={2020}
}

@INPROCEEDINGS{Once3D,
title={ONCE-3DLanes: Building Monocular 3D Lane Detection},
author= {Yan, Fan and Nie, Ming and Cai, Xinyue and Han, Jianhua and Xu, Hang and Yang, Zhen and Ye, Chaoqiang and Fu, Yanwei and Michael,Bi Mi and Zhang, Li},
booktitle={CVPR},
year={2022},
}

@INPROCEEDINGS {MapChangeDetection,
  author = {John Lambert and James Hays},
  title = {Trust, but Verify: Cross-Modality Fusion for {HD} Map Change Detection},
  booktitle = {NeurIPS},
  year = {2021}
}

@INPROCEEDINGS{A*3D,
  title={A* 3d dataset: Towards autonomous driving in challenging environments},
  author={Pham, Quang-Hieu and Sevestre, Pierre and Pahwa, Ramanpreet Singh and Zhan, Huijing and Pang, Chun Ho and Chen, Yuda and Mustafa, Armin and Chandrasekhar, Vijay and Lin, Jie},
  booktitle={ICRA},
  year={2020}
}

@ARTICLE{Boreas,
    author = {Keenan Burnett and David J Yoon and Yuchen Wu and Andrew Z Li and Haowei Zhang and Shichen Lu and Jingxing Qian and Wei-Kang Tseng and Andrew Lambert and Keith YK Leung and Angela P Schoellig and Timothy D Barfoot},
    title ={Boreas: A multi-season autonomous driving dataset},
    journal = {IJRR},
    year = {2023}
}

@ARTICLE{KITTI_360,
   title =  {{KITTI}-360: A Novel Dataset and Benchmarks for Urban Scene Understanding in 2D and 3D},
   author = {Yiyi Liao and Jun Xie and Andreas Geiger},
   journal = {TPAMI},
   year = {2022},
}

@INPROCEEDINGS{Ithaca365,
  title={Ithaca365: Dataset and driving perception under repeated and challenging weather conditions},
  author={Diaz-Ruiz, Carlos A and Xia, Youya and You, Yurong and Nino, Jose and Chen, Junan and Monica, Josephine and Chen, Xiangyu and Luo, Katie and Wang, Yan and Emond, Marc and others},
  booktitle={CVPR},
  year={2022}
}

@INPROCEEDINGS{CAM2BEV,
  author={Reiher, Lennart and Lampe, Bastian and Eckstein, Lutz},
  booktitle={ITSC}, 
  title={A Sim2Real Deep Learning Approach for the Transformation of Images from Multiple Vehicle-Mounted Cameras to a Semantically Segmented Image in Bird’s Eye View}, 
  year={2020}
}

@INPROCEEDINGS{aimotive,
  title={aimotive dataset: A multimodal dataset for robust autonomous driving with long-range perception},
  author={Matuszka, Tam{\'a}s and Barton, Iv{\'a}n and Butykai, {\'A}d{\'a}m and Hajas, P{\'e}ter and Kiss, D{\'a}vid and Kov{\'a}cs, Domonkos and Kuns{\'a}gi-M{\'a}t{\'e}, S{\'a}ndor and Lengyel, P{\'e}ter and N{\'e}meth, G{\'a}bor and Pet{\H{o}}, Levente and others},
  booktitle={ICLR Workshop Scene Representations for Autonomous Driving},
  year={2023}
}

@INPROCEEDINGS{cirrus,
  title = {Range adaptation for 3d object detection in lidar},
  author = {Wang, Ze and Ding, Sihao and Li, Ying and Zhao, Minming and Roychowdhury, Sohini and Wallin, Andreas and Sapiro, Guillermo and Qiu, Qiang},
  booktitle = {ICCV Workshops},
  year = {2019}
}

@ARTICLE{OpenMPD,
  author={Zhang, Xinyu and Li, Zhiwei and Gong, Yan and Jin, Dafeng and Li, Jun and Wang, Li and Zhu, Yanzhang and Liu, Huaping},
  journal={IEEE Transactions on Vehicular Technology}, 
  title={OpenMPD: An Open Multimodal Perception Dataset for Autonomous Driving}, 
  year={2022},
}

@ARTICLE{AIODrive,
author = {Weng, Xinshuo and Man, Yunze and Cheng, Dazhi and Park, Jinhyung and O'Toole, Matthew and Kitani, Kris},
journal = {arXiv},
title = {{All-In-One Drive: A Large-Scale Comprehensive Perception Dataset with High-Density Long-Range Point Clouds}},
year = {2020}
}

@inproceedings{Talk2Car, 
title={Talk2Car: Taking Control of Your Self-Driving Car}, 
author={Deruyttere, Thierry and Vandenhende, Simon and Grujicic, Dusan and Van Gool, Luc and Moens, Marie Francine}, 
booktitle={EMNLP IJCNLP}, 
pages={2088--2098}, 
year={2019}
}

@ARTICLE{pandaSet,
  author       = {Pengchuan Xiao and
                  Zhenlei Shao and
                  Steven Hao and
                  Zishuo Zhang and
                  Xiaolin Chai and
                  Judy Jiao and
                  Zesong Li and
                  Jian Wu and
                  Kai Sun and
                  Kun Jiang and
                  Yunlong Wang and
                  Diange Yang},
  title        = {PandaSet: Advanced Sensor Suite Dataset for Autonomous Driving},
  journal      = {arXiv preprint},
  volume       = {abs/2112.12610},
  year         = {2021}
}

@ARTICLE{CADC,
  author       = {Matthew Pitropov and
                  Danson Evan Garcia and
                  Jason Rebello and
                  Michael Smart and
                  Carlos Wang and
                  Krzysztof Czarnecki and
                  Steven Lake Waslander},
  title        = {Canadian Adverse Driving Conditions Dataset},
  journal      = {arXiv preprint},
  volume       = {abs/2001.10117},
  year         = {2020}
}

@INPROCEEDINGS{ONCE,
  author       = {Jiageng Mao and
                  Minzhe Niu and
                  Chenhan Jiang and
                  Hanxue Liang and
                  Xiaodan Liang and
                  Yamin Li and
                  Chaoqiang Ye and
                  Wei Zhang and
                  Zhenguo Li and
                  Jie Yu and
                  Hang Xu and
                  Chunjing Xu},
  title        = {One Million Scenes for Autonomous Driving: {ONCE} Dataset},
  year         = {2021},
  booktitle    = {NeurIPS Datasets and Benchmarks Track}
}

@INPROCEEDINGS{EgoSpeed,
  title={Ego Vehicle Speed Estimation using 3D Convolution with Masked Attention},
  author={Mathew, Athul M and Khalid, Thariq},
  booktitle={NeurIPS Track on Datasets and Benchmarks},
  year={2024}
}

@INPROCEEDINGS{Lyft,
  title={One thousand and one hours: Self-driving motion prediction dataset},
  author={Houston, John and Zuidhof, Guido and Bergamini, Luca and Ye, Yawei and Chen, Long and Jain, Ashesh and Omari, Sammy and Iglovikov, Vladimir and Ondruska, Peter},
  booktitle={CoRL},
  year={2021}
}

@ARTICLE{vod,
  author={Palffy, Andras and Pool, Ewoud and Baratam, Srimannarayana and Kooij, Julian F. P. and Gavrila, Dariu M.},
  journal={RA-L}, 
  title={Multi-Class Road User Detection With 3+1D Radar in the View-of-Delft Dataset}, 
  year={2022}
}

@INPROCEEDINGS{wei2024basal,
  title={{BaSAL}: Size-Balanced Warm Start Active Learning for LiDAR Semantic Segmentation},
  author={Wei, Jiarong and Lin, Yancong and Caesar, Holger},
  booktitle={ICRA},
  year={2024}
}

@INPROCEEDINGS{LidarMultiNet,
  title={Lidarmultinet: Towards a unified multi-task network for lidar perception},
  author={Ye, Dongqiangzi and Zhou, Zixiang and Chen, Weijia and Xie, Yufei and Wang, Yu and Wang, Panqu and Foroosh, Hassan},
  booktitle={AAAI},
  year={2023}
}

@ARTICLE{Cylinder3d,
  title={Cylinder3D: An Effective 3D Framework for Driving-scene LiDAR Semantic Segmentation},
  author={Zhou, Hui and Zhu, Xinge and Song, Xiao and Ma, Yuexin and Wang, Zhe and Li, Hongsheng and Lin, Dahua},
  journal={CVPR},
  year={2021}
}

@INPROCEEDINGS{SPVNAS,
  title={Searching efficient 3d architectures with sparse point-voxel convolution},
  author={Tang, Haotian and Liu, Zhijian and Zhao, Shengyu and Lin, Yujun and Lin, Ji and Wang, Hanrui and Han, Song},
  booktitle={ECCV},
  year={2020},
}

@ARTICLE{AMVNET,
  title={Amvnet: Assertion-based multi-view fusion network for lidar semantic segmentation},
  author={Liong, Venice Erin and Nguyen, Thi Ngoc Tho and Widjaja, Sergi and Sharma, Dhananjai and Chong, Zhuang Jie},
  journal={IJCAI},
  year={2021}
}

@INPROCEEDINGS{polarnet,
  title={Polarnet: An improved grid representation for online lidar point clouds semantic segmentation},
  author={Zhang, Yang and Zhou, Zixiang and David, Philip and Yue, Xiangyu and Xi, Zerong and Gong, Boqing and Foroosh, Hassan},
  booktitle={CVPR},
  year={2020}
}

@INPROCEEDINGS{rangenet++,
  title={Rangenet++: Fast and accurate lidar semantic segmentation},
  author={Milioto, Andres and Vizzo, Ignacio and Behley, Jens and Stachniss, Cyrill},
  booktitle={IROS},
  year={2019}
}

@INPROCEEDINGS{2D3DNet,
  title={Learning 3D semantic segmentation with only 2D image supervision},
  author={Genova, Kyle and Yin, Xiaoqi and Kundu, Abhijit and Pantofaru, Caroline and Cole, Forrester and Sud, Avneesh and Brewington, Brian and Shucker, Brian and Funkhouser, Thomas},
  booktitle={3DV},
  year={2021},
}

@INPROCEEDINGS{PHNet,
  title={Panoptic-PHNet: Towards Real-Time and High-Precision LiDAR Panoptic Segmentation via Clustering Pseudo Heatmap},
  author={Li, Jinke and He, Xiao and Wen, Yang and Gao, Yuan and Cheng, Xiaoqiang and Zhang, Dan},
  booktitle={CVPR},
  year={2022}
}

@ARTICLE{EfficientLPS,
  title={Efficientlps: Efficient lidar panoptic segmentation},
  author={Sirohi, Kshitij and Mohan, Rohit and B{\"u}scher, Daniel and Burgard, Wolfram and Valada, Abhinav},
  journal={IEEE Transactions on Robotics},
  year={2021},
}

@INPROCEEDINGS{PyOccNet,
  title={Predicting semantic map representations from images using pyramid occupancy networks},
  author={Roddick, Thomas and Cipolla, Roberto},
  booktitle={CVPR},
  year={2020}
}

@ARTICLE{VPN,
  title={Cross-view semantic segmentation for sensing surroundings},
  author={Pan, Bowen and Sun, Jiankai and Leung, Ho Yin Tiga and Andonian, Alex and Zhou, Bolei},
  journal={RA-L},
  year={2020}
}

@ARTICLE{OFT,  
  title={Orthographic feature transform for monocular 3d object detection},  
  author={Roddick, Thomas and Kendall, Alex and Cipolla, Roberto},  
  journal={BMVC},  
  year={2019}  
}

@INPROCEEDINGS{STA,
  title={Enabling spatio-temporal aggregation in birds-eye-view vehicle estimation},
  author={Saha, Avishkar and Mendez, Oscar and Russell, Chris and Bowden, Richard},
  booktitle={ICRA},
  year={2021}
}

@INPROCEEDINGS{Hft,
  title={{HFT}: Lifting perspective representations via hybrid feature transformation},
  author={Zou, Jiayu and Xiao, Junrui and Zhu, Zheng and Huang, Junjie and Huang, Guan and Du, Dalong and Wang, Xingang},
  booktitle={ICRA},
  year={2023}
}

@INPROCEEDINGS{Gitnet,
  title={Gitnet: Geometric prior-based transformation for birds-eye-view segmentation},
  author={Gong, Shi and Ye, Xiaoqing and Tan, Xiao and Wang, Jingdong and Ding, Errui and Zhou, Yu and Bai, Xiang},
  booktitle={ECCV},
  year={2022}
}

@ARTICLE{VED,
author = {Lu, Chenyang and van de Molengraft, Marinus Jacobus Gerardus and Dubbelman, Gijs},
journal = {RA-L},
title = {{Monocular Semantic Occupancy Grid Mapping With Convolutional Variational Encoder-Decoder Networks}},
volume = {4},
year = {2019}
}

@INPROCEEDINGS{Learning2look,
  title={Learning to look around objects for top-view representations of outdoor scenes},
  author={Schulter, Samuel and Zhai, Menghua and Jacobs, Nathan and Chandraker, Manmohan},
  booktitle={ECCV},
  year={2018}
}

@INPROCEEDINGS{Trans2maps,
  title={Translating images into maps},
  author={Saha, Avishkar and Mendez, Oscar and Russell, Chris and Bowden, Richard},
  booktitle={ICRA},
  year={2022}
}

@INPROCEEDINGS{EPOSH,
  title={Bird’s eye view segmentation using lifted 2D semantic features},
  author={Dwivedi, Isht and Malla, Srikanth and Chen, Yi-Ting and Dariush, Behzad},
  booktitle={BMVC},
  year={2021}
}

@INPROCEEDINGS{instagram,
  title={{InstaGraM}: Instance-level Graph Modeling for Vectorized {HD} Map Learning},
  author={Shin, Juyeb and Rameau, Francois and Jeong, Hyeonjun and Kum, Dongsuk},
  booktitle={CVPR VCAD Workshop},
  year={2023}
}

@INPROCEEDINGS{vectormapnet,
  title={Vectormapnet: End-to-end vectorized {HD} map learning},
  author={Liu, Yicheng and Wang, Yue and Wang, Yilun and Zhao, Hang},
  booktitle={ICML},
  year={2023}
}

@INPROCEEDINGS{hdmapnet,
  title={{HDMapNet}: An online {HD} map construction and evaluation framework},
  author={Li, Qi and Wang, Yue and Wang, Yilun and Zhao, Hang},
  booktitle={ICRA},
  year={2022}
}

@INPROCEEDINGS{bevsegformer,
  title={{BEVSegFormer}: Bird's Eye View Semantic Segmentation From Arbitrary Camera Rigs},
  author={Peng, Lang and Chen, Zhirong and Fu, Zhangjie and Liang, Pengpeng and Cheng, Erkang},
  booktitle={WACV},
  year={2023}
}

@ARTICLE{boschstreet,
  title={Bosch Street Dataset: A Multi-Modal Dataset with Imaging Radar for Automated Driving},
  author={Armanious, Karim and Quach, Maurice and Ulrich, Michael and Winterling, Timo and Friesen, Johannes and Braun, Sascha and Jenet, Daniel and Feldman, Yuri and Kosman, Eitan and Rapp, Philipp and others},
  journal={arXiv preprint arXiv:2407.12803},
  year={2024}
}

@ARTICLE{a2d2,
  title={{A2D2}: Audi autonomous driving dataset},
  author={Geyer, Jakob and Kassahun, Yohannes and Mahmudi, Mentar and Ricou, Xavier and Durgesh, Rupesh and Chung, Andrew S and Hauswald, Lorenz and Pham, Viet Hoang and M{\"u}hlegg, Maximilian and Dorn, Sebastian and others},
  journal={arXiv preprint arXiv:2004.06320},
  year={2020}
}

@INPROCEEDINGS{liftsplatshoot,
  title={Lift, splat, shoot: Encoding images from arbitrary camera rigs by implicitly unprojecting to 3d},
  author={Philion, Jonah and Fidler, Sanja},
  booktitle={ECCV},
  year={2020}
}

@INPROCEEDINGS{PYVA,
  title={Projecting your view attentively: Monocular road scene layout estimation via cross-view transformation},
  author={Yang, Weixiang and Li, Qi and Liu, Wenxi and Yu, Yuanlong and Ma, Yuexin and He, Shengfeng and Pan, Jia},
  booktitle={CVPR},
  year={2021}
}

@INPROCEEDINGS{caesar2020nuscenes,
  title={{nuScenes}: A multimodal dataset for autonomous driving},
  author={Caesar, Holger and Bankiti, Varun and Lang, Alex H. and Vora, Sourabh and Liong, Venice Erin and Xu, Qiang and Krishnan, Anush and Pan, Yu and Baldan, Giancarlo and Beijbom, Oscar},
  booktitle={CVPR},
  year={2020}
}

@INPROCEEDINGS{fong2022nuscenes-panoptic,
  title={Panoptic nuScenes: A Large-Scale Benchmark for LiDAR Panoptic Segmentation and Tracking},
  author={Fong, Whye Kit and Mohan, Rohit and Hurtado, Juana and Zhou, Lubing and Caesar, Holger and Beijbom, Oscar and Valada, Abhinav}, 
  booktitle={ICRA},
  year=2022
}

@INPROCEEDINGS{guo2020nuscenes-pkl,
  title={The efficacy of Neural Planning Metrics: A meta-analysis of PKL on nuScenes},
  author={Guo, Yiluan and Caesar, Holger and Beijbom, Oscar and Philion, Jonah and Fidler, Sanja},
  booktitle={IROS 2020 Workshop on Benchmarking Progress in Autonomous Driving},
  year=2020
}

@INPROCEEDINGS{PointPainting,
  title={Pointpainting: Sequential fusion for 3d object detection},
  author={Vora, Sourabh and Lang, Alex H and Helou, Bassam and Beijbom, Oscar},
  booktitle={CVPR},
  year={2020}
}

@inproceedings{zhang2024sparselif,
  title={SparseLIF: High-performance sparse LiDAR-camera fusion for 3D object detection},
  author={Zhang, Hongcheng and Liang, Liu and Zeng, Pengxin and Song, Xiao and Wang, Zhe},
  booktitle={ECCV},
  pages={109--128},
  year={2024},
  organization={Springer}
}

@article{lin2022sparse4d,
  title={Sparse4d: Multi-view 3d object detection with sparse spatial-temporal fusion},
  author={Lin, Xuewu and Lin, Tianwei and Pei, Zixiang and Huang, Lichao and Su, Zhizhong},
  journal={arXiv preprint arXiv:2211.10581},
  year={2022}
}

@article{lin2023sparse4d,
  title={Sparse4d v2: Recurrent temporal fusion with sparse model},
  author={Lin, Xuewu and Lin, Tianwei and Pei, Zixiang and Huang, Lichao and Su, Zhizhong},
  journal={arXiv preprint arXiv:2305.14018},
  year={2023}
}

@inproceedings{yogamani2019woodscape,
  title={Woodscape: A multi-task, multi-camera fisheye dataset for autonomous driving},
  author={Yogamani, Senthil and Hughes, Ciar{\'a}n and Horgan, Jonathan and Sistu, Ganesh and Varley, Padraig and O'Dea, Derek and Uric{\'a}r, Michal and Milz, Stefan and Simon, Martin and Amende, Karl and others},
  booktitle={ICCV},
  pages={9308--9318},
  year={2019}
}

@misc{mmdet3d2020,
    title={{MMDetection3D: OpenMMLab} next-generation platform for general {3D} object detection},
    author={MMDetection3D Contributors},
    year={2020}
}

@ARTICLE{grisetti2010graph-based-slam,
  author={Grisetti, Giorgio and Kümmerle, Rainer and Stachniss, Cyrill and Burgard, Wolfram},
  journal={IEEE ITS Magazine}, 
  title={A Tutorial on Graph-Based SLAM}, 
  year={2010}
}

@ARTICLE{ICP,
  author={Besl, P.J. and McKay, Neil D.},
  journal={TPAMI}, 
  title={A method for registration of 3-D shapes}, 
  year={1992},
  volume={14},
  number={2}
}

@ARTICLE{LoopClosureSurvey,
  title={Role of deep learning in loop closure detection for visual and lidar slam: A survey},
  author={Arshad, Saba and Kim, Gon-Woo},
  journal={Sensors},
  year={2021}
}

@ARTICLE{rethinkingopenloop,
  title={Rethinking the Open-Loop Evaluation of End-to-End Autonomous Driving in nuScenes},
  author={Zhai, Jiang-Tian and Feng, Ze and Du, Jinhao and Mao, Yongqiang and Liu, Jiang-Jiang and Tan, Zichang and Zhang, Yifu and Ye, Xiaoqing and Wang, Jingdong},
  journal={arXiv preprint arXiv:2305.10430},
  year={2023}
}

@inproceedings{hu2022st,
  title={St-p3: End-to-end vision-based autonomous driving via spatial-temporal feature learning},
  author={Hu, Shengchao and Chen, Li and Wu, Penghao and Li, Hongyang and Yan, Junchi and Tao, Dacheng},
  booktitle={ECCV},
  pages={533--549},
  year={2022},
  organization={Springer}
}

@inproceedings{hu2023planning,
  title={Planning-oriented autonomous driving},
  author={Hu, Yihan and Yang, Jiazhi and Chen, Li and Li, Keyu and Sima, Chonghao and Zhu, Xizhou and Chai, Siqi and Du, Senyao and Lin, Tianwei and Wang, Wenhai and others},
  booktitle={CVPR},
  pages={17853--17862},
  year={2023}
}

@inproceedings{jiang2023vad,
  title={Vad: Vectorized scene representation for efficient autonomous driving},
  author={Jiang, Bo and Chen, Shaoyu and Xu, Qing and Liao, Bencheng and Chen, Jiajie and Zhou, Helong and Zhang, Qian and Liu, Wenyu and Huang, Chang and Wang, Xinggang},
  booktitle={ICCV},
  year={2023}
}

@ARTICLE{li2023toponet,
  title={Topology Reasoning for Driving Scenes},
  author={Li, Tianyu and Chen, Li and Geng, Xiangwei and Wang, Huijie and Li, Yang and Liu, Zhenbo and Jiang, Shengyin and Wang, Yuting and Xu, Hang and Xu, Chunjing and Wen, Feng and Luo, Ping and Yan, Junchi and Zhang, Wei and Wang, Xiaogang and Qiao, Yu and Li, Hongyang},
  journal={arXiv preprint arXiv:2304.05277},
  year={2023}
}

@ARTICLE{huang2018apolloscape,
    title={The ApolloScape Dataset for Autonomous Driving},
    author={Xinyu, Huang and Xinjing, Cheng and Qichuan, Geng and Binbin, Cao and Dingfu, Zhou and Peng, Wang and Yuanqing, Lin and Ruigang, Yang},
    journal={arXiv preprint},
    year={2018},
}

@INPROCEEDINGS{reason2drive,
      title={Reason2Drive: Towards Interpretable and Chain-based Reasoning for Autonomous Driving}, 
      author={Ming Nie and Renyuan Peng and Chunwei Wang and Xinyue Cai and Jianhua Han and Hang Xu and Li Zhang},
      booktitle={ECCV},
      year={2024}
}

@INPROCEEDINGS{wu2023referringmultiobjecttracking,
      title={Referring Multi-Object Tracking}, 
      author={Dongming Wu and Wencheng Han and Tiancai Wang and Xingping Dong and Xiangyu Zhang and Jianbing Shen},
      booktitle={CVPR},
      year={2023}
}

@misc{schreier2023offline,
    title={On Offline Evaluation of 3D Object Detection for Autonomous Driving}, 
    author={Tim Schreier and Katrin Renz and Andreas Geiger and Kashyap Chitta},
    year={2023},
    eprint={2308.12779},
    archivePrefix={arXiv},
    primaryClass={cs.CV}
}

@INPROCEEDINGS{neural_map_planner,
  author  = {Xiong, Xuan and Liu, Yicheng and Yuan, Tianyuan and Wang, Yue and Wang, Yilun and Zhao Hang},
  title   = {Neural Map Prior for Autonomous Driving},
  year    = {2023},
  booktitle={CVPR},
}

@INPROCEEDINGS{leveraging_trajectory_prediction,
  title={Leveraging Future Relationship Reasoning for Vehicle Trajectory Prediction},
  author={Park, Daehee and Ryu, Hobin and Yang, Yunseo and Cho, Jegyeong and Kim, Jiwon and Yoon, Kuk-Jin},
  booktitle={ICLR},
  year={2023}
}

@INPROCEEDINGS{hdnet,
  title={Hdnet: Exploiting {HD} maps for 3d object detection},
  author={Yang, Bin and Liang, Ming and Urtasun, Raquel},
  booktitle={CoRL},
  year={2018}
}

@INPROCEEDINGS{mp3,
  title={Mp3: A unified model to map, perceive, predict and plan},
  author={Casas, Sergio and Sadat, Abbas and Urtasun, Raquel},
  booktitle={CVPR},
  year={2021}
}

@ARTICLE{Lane1,
  title={Lane graph estimation for scene understanding in urban driving},
  author={Z{\"u}rn, Jannik and Vertens, Johan and Burgard, Wolfram},
  journal={RA-L},
  year={2021}
}

@ARTICLE{Lane2,
  title={Learning to Predict Navigational Patterns from Partial Observations},
  author={Karlsson, Robin and Carballo, Alexander and Lepe-Salazar, Francisco and Fujii, Keisuke and Ohtani, Kento and Takeda, Kazuya},
  journal={RA-L},
  year={2023}
}

@INPROCEEDINGS{khurana2023point,
  title={Point Cloud Forecasting as a Proxy for 4D Occupancy Forecasting},
  author={Khurana, Tarasha and Hu, Peiyun and Held, David and Ramanan, Deva},
  booktitle={CVPR},
  year={2023},
}

@INPROCEEDINGS{yan2022once,
 title={ONCE-3DLanes: Building Monocular 3D Lane Detection},
 author= {Yan, Fan and Nie, Ming and Cai, Xinyue and Han, Jianhua and Xu, Hang and Yang, Zhen and Ye, Chaoqiang and Fu, Yanwei and Michael,Bi Mi and Zhang, Li},
 booktitle={CVPR},
 year={2022},
}

@ARTICLE{VOD-prediction,
  author={Boekema, Hidde J-H. and Martens, Bruno K.W. and Kooij, Julian F.P. and Gavrila, Dariu M.},
  journal={RA-L}, 
  title={Multi-Class Trajectory Prediction in Urban Traffic Using the View-of-Delft Prediction Dataset}, 
  year={2024}
}

@INPROCEEDINGS{semantic_kitti,
  author = {J. Behley and M. Garbade and A. Milioto and J. Quenzel and S. Behnke and C. Stachniss and J. Gall},
  title = {{SemanticKITTI: A Dataset for Semantic Scene Understanding of LiDAR Sequences}},
  booktitle = {ICCV},
  year = {2019}
}

@INPROCEEDINGS{geiger2012kitti,
    title={{Are we ready for Autonomous Driving? The KITTI Vision Benchmark Suite}},
    author={A. Geiger and P. Lenz and R. Urtasun},
    journal={CVPR},
    year={2012},
    booktitle={CVPR},
}

@ARTICLE{sun2019wod,
    author={Pei, Sun and Henrik, Kretzschmar and Xerxes, Dotiwalla and Aurelien, Chouard and Vijaysai, Patnaik and Paul, Tsui and James, Guo and Yin, Zhou and Yuning, Chai and Benjamin, Caine and Vijay, Vasudevan and Wei, Han and Jiquan, Ngiam and Hang, Zhao and Aleksei, Timofeev and Scott, Ettinger and Maxim, Krivokon and Amy, Gao and Aditya, Joshi and Yu, Zhang and Jonathon, Shlens and Zhifeng, Chen and Dragomir, Anguelov},
    title={Scalability in Perception for Autonomous Driving: Waymo Open Dataset},
    journal={CVPR},
    year={2019},
}

@INPROCEEDINGS{deruyttere2019talk2car,
  title={Talk2Car: Taking Control of Your Self-Driving Car},
  author={Deruyttere, Thierry and Vandenhende, Simon and Grujicic, Dusan and Van Gool, Luc and Moens, Marie Francine},
  booktitle={EMNLP-IJCNLP},
  year={2019}
}

@ARTICLE{qian2023nuscenesqa,
  title={NuScenes-QA: A Multi-modal Visual Question Answering Benchmark for Autonomous Driving Scenario},
  author={Qian, Tianwen and Chen, Jingjing and Zhuo, Linhai and Jiao, Yang and Jiang, Yu-Gang},
  journal={arXiv preprint arXiv:2305.14836},
  year={2023}
}

@ARTICLE{wu2023nuprompt,
  title={Language Prompt for Autonomous Driving},
  author={Wu, Dongming and Han, Wencheng and Wang, Tiancai and Liu, Yingfei and Zhang, Xiangyu and Shen, Jianbing},
  journal={arXiv preprint},
  year={2023}
}

@ARTICLE{zhang2023motiontrack,
    title={MotionTrack: End-to-End Transformer-based Multi-Object Tracing with LiDAR-Camera Fusion}, 
    author={Ce, Zhang and Chengjie, Zhang and Yiluan, Guo and Lingji, Chen and Michael, Happold},
    journal={arXiv preprint},
    year={2023},
}

@INPROCEEDINGS{ruppel2022transmot,
   title={Transformers for Multi-Object Tracking on Point Clouds},
   author={Ruppel, Felicia and Faion, Florian and Glaser, Claudius and Dietmayer, Klaus},
   publisher={IEEE},
   year={2022},
   booktitle={IV}, 
}

@misc{kirillov2019panoptic,
    title={Panoptic Segmentation}, 
    author={Alexander Kirillov and Kaiming He and Ross Girshick and Carsten Rother and Piotr Dollár},
    year={2019},
    journal={CVPR},
}

@INPROCEEDINGS{li2023egostatus,
    title={Is Ego Status All You Need for Open-Loop End-to-End Autonomous Driving?}, 
    author={Zhiqi Li and Zhiding Yu and Shiyi Lan and Jiahan Li and Jan Kautz and Tong Lu and Jose M. Alvarez},
    year={2024},
    booktitle={CVPR}
}

@ARTICLE{caesar2022nuplan,
    title={{nuPlan}: A closed-loop ML-based planning benchmark for autonomous vehicles}, 
    author={Holger Caesar and Juraj Kabzan and Kok Seang Tan and Whye Kit Fong and Eric Wolff and Alex Lang and Luke Fletcher and Oscar Beijbom and Sammy Omari},
    year={2022},
    eprint={2106.11810},
    archivePrefix={arXiv},
    journal={arXiv preprint}
}

@ARTICLE{karnchanachari24nuplan,
    author={Napat Karnchanachari and Dimitris Geromichalos and Kok Seang Tan and Nanxiang Li and Christopher Eriksen and Shakiba Yaghoubi and Noushin Mehdipour and Gianmarco Bernasconi and Whye Kit Fong and Yiluan Guo and Holger Caesar},
    title={Towards learning-based planning: The {nuPlan} benchmark for real-world autonomous driving},
    journal={ICRA},
    year={2024},
}

@INPROCEEDINGS{tian2023occ3d,
  title={Occ3D: A Large-Scale 3D Occupancy Prediction Benchmark for Autonomous Driving},
  author={Tian, Xiaoyu and Jiang, Tao and Yun, Longfei and Wang, Yue and Wang, Yilun and Zhao, Hang},
  booktitle={NeurIPS},
  year={2023},
}

@ARTICLE{sima2023occnet,
    title={Scene as Occupancy},
    author={Chonghao Sima and Wenwen Tong and Tai Wang and Li Chen and Silei Wu and Hanming Deng and Yi Gu and Lewei Lu and Ping Luo and Dahua Lin and Hongyang Li},
    year={2023},
    eprint={2306.02851},
    journal={arXiv preprint}
}

@INPROCEEDINGS{frome2009streetview,
  author={Frome, Andrea and Cheung, German and Abdulkader, Ahmad and Zennaro, Marco and Wu, Bo and Bissacco, Alessandro and Adam, Hartwig and Neven, Hartmut and Vincent, Luc},
  booktitle={ICCV}, 
  title={Large-scale privacy protection in Google Street View}, 
  year={2009}
}

@INPROCEEDINGS{caesar2016cocostuff,
  author       = {Holger Caesar and
                  Jasper R. R. Uijlings and
                  Vittorio Ferrari},
  booktitle={CVPR}, 
  title        = {{COCO-Stuff}: Thing and Stuff Classes in Context},
  year         = {2016},
}

@INPROCEEDINGS{yin2021center,
  title={Center-based 3D Object Detection and Tracking},
  author={Yin, Tianwei and Zhou, Xingyi and Kr{\"a}henb{\"u}hl, Philipp},
  booktitle={CVPR},
  year={2021},
}

@INPROCEEDINGS{nabati2020centerfusion,
  title={CenterFusion: Center-based Radar and Camera Fusion for 3D Object Detection},
  author={Nabati, Ramin and Qi, Hairong},
  booktitle={WACV},
  year={2021}
}

@INPROCEEDINGS{Yang2022BEVFormerV2,
  title={BEVFormer v2: Adapting Modern Image Backbones to Bird's-Eye-View Recognition via Perspective Supervision},
  author={Chenyu Yang and Yuntao Chen and Haofei Tian and Chenxin Tao and Xizhou Zhu and Zhaoxiang Zhang and Gao Huang and Hongyang Li and Y. Qiao and Lewei Lu and Jie Zhou and Jifeng Dai},
  booktitle={CVPR},
  year={2023},
}

@ARTICLE{Weng2020AB3DMOT, 
author = {Weng, Xinshuo and Wang, Jianren and Held, David and Kitani, Kris}, 
journal = {IROS}, 
title = {{3D Multi-Object Tracking: A Baseline and New Evaluation Metrics}}, 
year = {2020} 
}

@INPROCEEDINGS{liu2023fullysparse3dpanopticoccupancyprediction,
  title={Fully sparse 3d panoptic occupancy prediction},
  author={Liu, Haisong and Wang, Haiguang and Chen, Yang and Yang, Zetong and Zeng, Jia and Chen, Li and Wang, Limin},
  booktitle={ECCV},
  year={2024}
}

@ARTICLE{luiten2020hota,
   title={HOTA: A Higher Order Metric for Evaluating Multi-object Tracking},
   volume={129},
   number={2},
   journal={IJCV},
   author={Luiten, Jonathon and Osep, Aljos̆a and Dendorfer, Patrick and Torr, Philip and Geiger, Andreas and Leal-Taixé, Laura and Leibe, Bastian},
   year={2020}
}

@INPROCEEDINGS{argoverse2,
  author = {Benjamin Wilson and William Qi and Tanmay Agarwal and John Lambert and Jagjeet Singh and Siddhesh Khandelwal and Bowen Pan and Ratnesh Kumar and Andrew Hartnett and Jhony Kaesemodel Pontes and Deva Ramanan and Peter Carr and James Hays},
  title = {Argoverse 2: Next Generation Datasets for Self-Driving Perception and Forecasting},
  booktitle = {NeurIPS Datasets and Benchmarks},
  year = {2021}
}

@INPROCEEDINGS{liu2023offlinetrackingwithobjectpermanence,
      title={Offline Tracking with Object Permanence}, 
      author={Xianzhong Liu and Holger Caesar},
      booktitle={IV},
      year={2024}
}

@INPROCEEDINGS{lilja2024dataleakage,
  author = {Adam Lilja and Junsheng Fu and Erik Stenborg and Lars Hammarstrand},
  title = {Localization Is All You Evaluate: Data Leakage in Online Mapping Datasets and How to Fix It},
  booktitle = {CVPR},
  year = {2024}
}

@INPROCEEDINGS{truckscenes2024,
 title = {MAN TruckScenes: A multimodal dataset for autonomous trucking in diverse conditions},
 author = {Fent, Felix and Kuttenreich, Fabian and Ruch, Florian and Rizwin, Farija and Juergens, Stefan and Lechermann, Lorenz and Nissler, Christian and Perl, Andrea and Voll, Ulrich and Yan, Min and Lienkamp, Markus},
 booktitle = {NeurIPS},
 editor = {A. Globerson and L. Mackey and D. Belgrave and A. Fan and U. Paquet and J. Tomczak and C. Zhang},
 year = {2024}
}

@INPROCEEDINGS{tj4dradset,  
author={Zheng, Lianqing and Ma, Zhixiong and Zhu, Xichan and Tan, Bin and Li, Sen and Long, Kai and Sun, Weiqi and Chen, Sihan and Zhang, Lu and Wan, Mengyue and Huang, Libo and Bai, Jie},  
booktitle={ITSC},   
title={TJ4DRadSet: A 4D Radar Dataset for Autonomous Driving},   
year={2022}
}

@INPROCEEDINGS{zendardataset,
  title={High-resolution radar dataset for semi-supervised learning of dynamic objects},
  author={Mostajabi, Mohammadreza and Wang, Ching Ming and Ranjan, Darsh and Hsyu, Gilbert},
  booktitle={CVPR Workshops},
  year={2020}
}

@INPROCEEDINGS{pklmetric,
  title={Learning to Evaluate Perception Models using Planner-Centric Metrics},
  author={Philion, Jonah and Kar, Amlan and Fidler, Sanja},
  booktitle={CVPR},
  year={2020}
}

@INPROCEEDINGS{qi2021offboardperception,
  title={Offboard 3D Object Detection from Point Cloud Sequences},
  author={Charles R. Qi and Yin Zhou and Mahyar Najibi and Pei Sun and Khoa Vo and Boyang Deng and Dragomir Anguelov},
  booktitle={CVPR},
  year={2021}
}

@INPROCEEDINGS{drivelm,
  title={{DriveLM}: Driving with Graph Visual Question Answering},
  author={Chonghao Sima and Katrin Renz and Kashyap Chitta and Li Chen and Hanxue Zhang and Chengen Xie and Jens Beißwenger and Ping Luo and Andreas Geiger and Hongyang Li},
  booktitle={ECCV},
  year={2024}
}

@article{nuscenesspatialqa,
    title={{NuScenes-SpatialQA}: A Spatial Understanding and Reasoning Benchmark for Vision-Language Models in Autonomous Driving},
    author={Tian, Kexin and Mao, Jingrui and Zhang, Yunlong and Jiang, Jiwan and Zhou, Yang and Tu, Zhengzhong},
    journal={arXiv preprint arXiv:2504.03164},
    year={2025}
}

@InProceedings{nuscenesmqa,
    author    = {Inoue, Yuichi and Yada, Yuki and Tanahashi, Kotaro and Yamaguchi, Yu},
    title     = {NuScenes-MQA: Integrated Evaluation of Captions and QA for Autonomous Driving Datasets Using Markup Annotations},
    booktitle = {WACV Workshops},
    year      = {2024}
}

@inproceedings{omnidrive,
  title={{OmniDrive}: A Holistic Vision-Language Dataset for Autonomous Driving with Counterfactual Reasoning},
  author={Shihao Wang and Zhiding Yu and Xiaohui Jiang and Shiyi Lan and Min Shi and Nadine Chang and Jan Kautz and Ying Li and Jose M. Alvarez},
  booktitle={CVPR},
  year={2025}
}

@misc{comma-2k19,
Author = {Harald Schafer and Eder Santana and Andrew Haden and Riccardo Biasini},
Title = {A Commute in Data: The comma2k19 Dataset},
Year = {2018},
Eprint = {arXiv:1812.05752},
}

@article{lei2023hvdetfusion,
  title={Hvdetfusion: A simple and robust camera-radar fusion framework},
  author={Lei, Kai and Chen, Zhan and Jia, Shuman and Zhang, Xiaoteng},
  journal={arXiv preprint arXiv:2307.11323},
  year={2023}
}

@INPROCEEDINGS{hung2024,
  author={Hung, Wei-Chih and Casser, Vincent and Kretzschmar, Henrik and Hwang, Jyh-Jing and Anguelov, Dragomir},
  booktitle={ICRA}, 
  title={LET-3D-AP: Longitudinal Error Tolerant 3D Average Precision for Camera-Only 3D Detection}, 
  year={2024}
}

@INPROCEEDINGS{nuscenes-misalignment,
author = {Yang, Wei-Jong and Ho, Li-Yang},
booktitle = { 2023 IEEE/ACIS 8th International Conference on Big Data, Cloud Computing, and Data Science (BCD)},
title = {{ Addressing Data Misalignment in Image-LiDAR Fusion on Point Cloud Segmentation }},
year = {2023},
volume = {},
ISSN = {},
pages = {365-367}
}

@InProceedings{radial,
author = {Rebut, Julien and Ouaknine, Arthur and Malik, Waqas and P\'erez, Patrick},
title = {Raw High-Definition Radar for Multi-Task Learning},
booktitle = {CVPR},
year = {2022}
}
}

\vspace{-8mm}

\begin{IEEEbiography}[{\includegraphics[width=1.12in,clip,keepaspectratio]{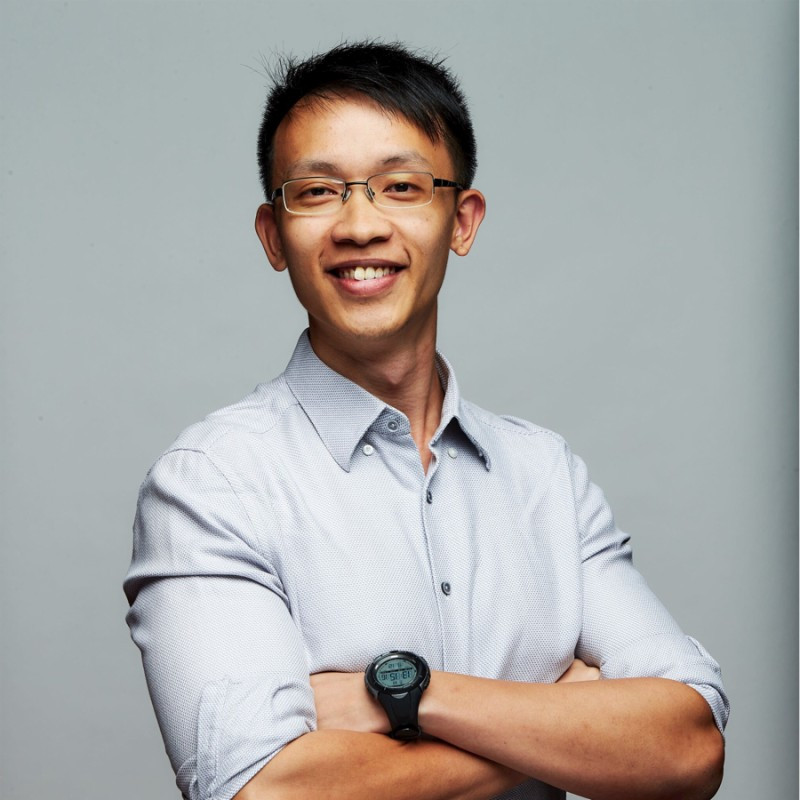}}]{Whye Kit Fong}
Whye Kit Fong is a Senior Research Engineer whose expertise and interest are in applied machine learning. He is currently at Motional (formerly nuTonomy), where he has worked on a broad spectrum of projects spanning from designing and training models for autonomous vehicles, to mining of large-scale datasets, to ensuring that models are scalable and production-ready. During his time at Motional, he was also responsible for the release of Panoptic nuScenes and contributed to nuPlan.
\end{IEEEbiography}

\vspace{-9mm}

\begin{IEEEbiography}[{\includegraphics[width=1.12in,clip,keepaspectratio]{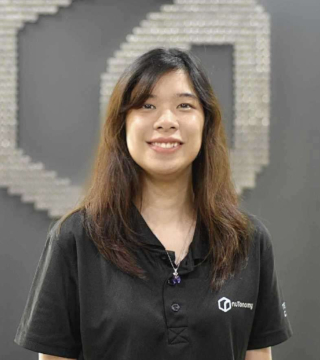}}]{Venice Erin Liong}
Dr. Venice Erin Liong is currently a Principal Research Scientist and Technical Lead Manager at Motional (formerly nuTonomy) in the Singapore branch. She has more than seven years of industry experience in autonomous vehicle perception. In particular, she has worked on several perception tasks such as 2D object detection, 3D object detection, point-based segmentation and map element detection using various sensor modalities and fusion techniques. She and her team have experience on deploying deep learning solutions on-car for on-road performance and off-car for scalable automated annotations. 
\end{IEEEbiography}

\vspace{-9mm}

\begin{IEEEbiography}[{\includegraphics[width=1.12in,clip,keepaspectratio]{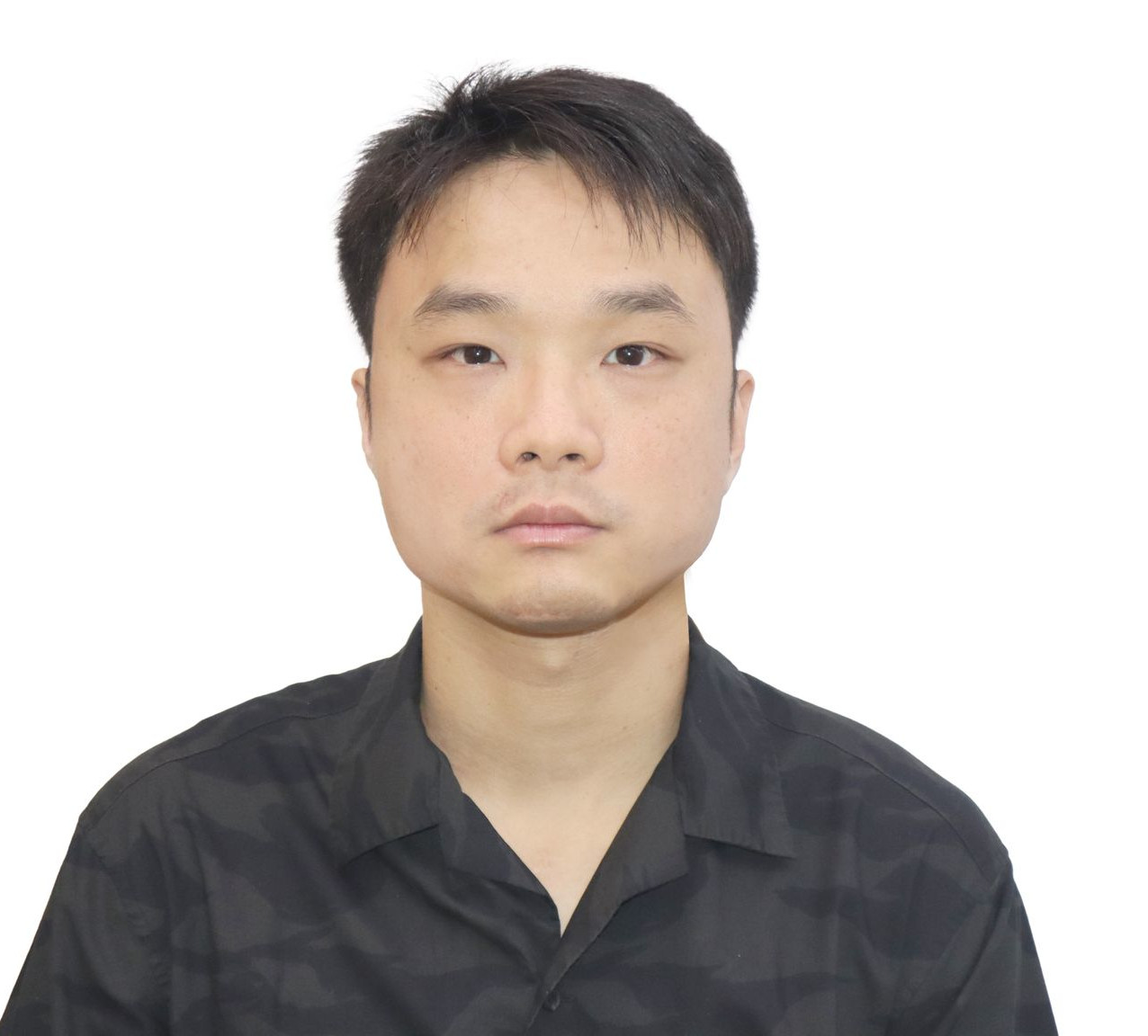}}]{Kok Seang Tan}
Kok Seang Tan was a research engineer on the Data Mining team at Motional (formerly nuTonomy). He is currently working as a Perception Engineer at TIER IV, focusing on data efficiency in perception. His research interests are primarily in autonomous vehicles, with a specific focus on data efficiency, perception, and prediction. He has also contributed to the nuPlan dataset.
\end{IEEEbiography}

\vspace{-5mm}

\begin{IEEEbiography}[{\includegraphics[width=1.12in,clip,keepaspectratio]{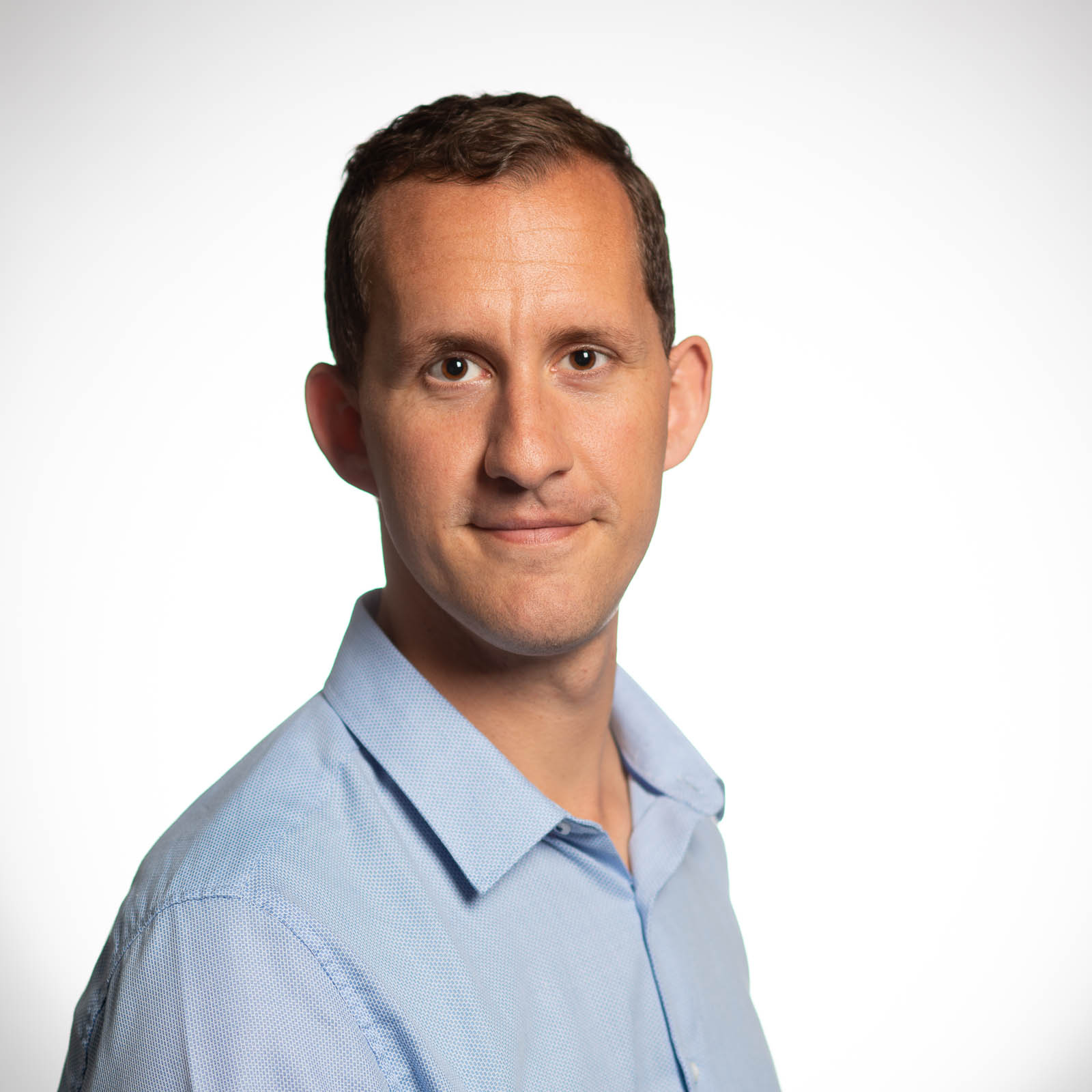}}]{Holger Caesar}
Dr. Holger Caesar is a tenured Assistant Professor in the Intelligent Vehicles group of TU Delft in the Netherlands. Holger's research interests are in the area of Autonomous Vehicle perception and prediction, with a particular focus on scalability of learning and annotation approaches. Previously Holger was a Principal Research Scientist at Motional (formerly nuTonomy). He is best known for leading the influential autonomous driving datasets nuScenes and nuPlan, as well as his contributions to the 3d object detection method PointPillars and the photorealistic crash-test Neuro NCAP.
\end{IEEEbiography}

\vfill

\end{document}